%% file: main.tex
\documentclass[10pt,twocolumn,letterpaper]{article}

\usepackage{cvpr}              

\input{preamble}

\usepackage{graphicx}
\usepackage{amsmath}
\usepackage{amssymb}
\usepackage{booktabs}
\usepackage{gensymb}
\usepackage{multirow}
\usepackage{rotating}
\usepackage{fancyhdr}
\usepackage{latexsym}
\usepackage{url}
\usepackage{algorithm}
\usepackage{algorithmic}
\usepackage{bm}
\usepackage{stmaryrd}
\usepackage{dsfont}
\usepackage{epstopdf}
\usepackage{booktabs}
\usepackage{dashrule}
\usepackage{soul}
\usepackage{color, colortbl}
\usepackage{xcolor}
\usepackage[accsupp]{axessibility} 

\usepackage{pifont}
\usepackage{enumitem}
\usepackage{tcolorbox}
\definecolor{Gray}{gray}{0.9}
\definecolor{demphcolor}{RGB}{144,144,144}
\definecolor{backblue}{RGB}{221,239,251}
\definecolor{backblue1}{RGB}{186,216,242}
\definecolor{backred}{RGB}{244,199,204}

\newcommand{\demph}[1]{\textcolor{demphcolor}{#1}}

\newcommand\blfootnote[1]{
    \begingroup
    \renewcommand\thefootnote{}\footnote{#1}
    \addtocounter{footnote}{-1}
    \endgroup
}

\definecolor{cvprblue}{rgb}{0.21,0.49,0.74}
\usepackage[pagebackref,breaklinks,colorlinks,citecolor=cvprblue]{hyperref}
\newcommand{\method}{\textsc{EVCap}}

\def\paperID{6846} 
\def\confName{CVPR}
\def\confYear{2024}

\title{\method: Retrieval-Augmented Image Captioning \\ with External Visual--Name Memory for Open-World Comprehension}

\author{Jiaxuan Li$^{1*}$,\ \ Duc Minh Vo$^{1*}$,\ \ Akihiro Sugimoto$^{2}$,\ \ Hideki Nakayama$^{1}$\\
$^{1}$The University of Tokyo, Japan\ \ $^{2}$National Institute of Informatics, Japan \\
\tt\small \{li,vmduc\}@nlab.ci.i.u-tokyo.ac.jp\ \ sugimoto@nii.ac.jp\ \ nakayama@ci.i.u-tokyo.ac.jp\\
}

\begin{document}

\addtolength{\baselineskip}{-0.08pt}

\maketitle

\begin{abstract}

Large language models (LLMs)-based image captioning has the capability of describing objects not explicitly observed in training data; yet novel objects occur frequently, necessitating the requirement of sustaining up-to-date object knowledge for open-world comprehension.
Instead of relying on large amounts of data and/or scaling up network parameters, we introduce a highly effective retrieval-augmented image captioning method that prompts LLMs with object names retrieved from \textbf{E}xternal \textbf{V}isual--name memory (\method).
We build ever-changing object knowledge memory using objects' visuals and names, enabling us to (i) update the memory at a minimal cost and (ii) effortlessly augment LLMs with retrieved object names by utilizing a lightweight and fast-to-train model.
Our model, which was trained only on the COCO dataset, can adapt to out-of-domain without requiring additional fine-tuning or re-training. 
Our experiments conducted on benchmarks and synthetic commonsense-violating data show that \method{}, with only 3.97M trainable parameters, 
exhibits superior performance compared to other methods based on frozen pre-trained LLMs.
Its performance is also competitive to specialist SOTAs that require extensive training.
\blfootnote{\textsuperscript{*Equal contributions. Code is available at \url{https://jiaxuan-li.github.io/EVCap}.}}

\end{abstract}

\section{Introduction}

Advanced image captioning based on large language models (LLMs)~\cite{alayrac2022flamingo, chen2023pali,li2023blip2, chen2023palix} has focused on the approach using big-scale models trained on ever-increasingly large-scale datasets, which is no longer viable. 
This is because the computational cost 
to train the models
increases exponentially and, more importantly, updating training data is almost impossible to keep pace with the growth of novel objects in our daily lives. 
Sustaining ever-changing object knowledge with a reasonable cost is a pressing concern in LLMs-based models to truly unlock open-world comprehension.

\begin{figure}[tb]
\centering\includegraphics[width=0.47
\textwidth]{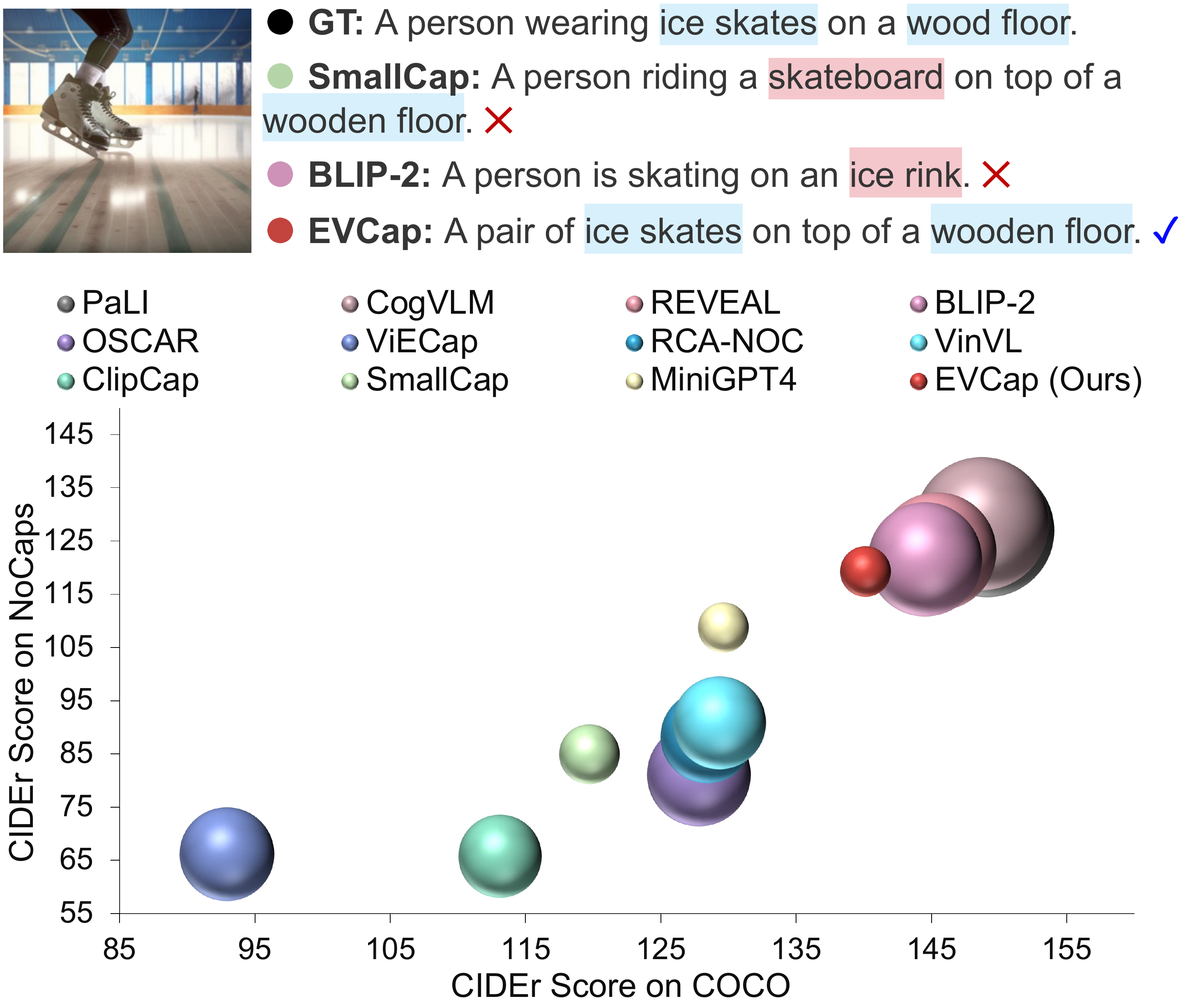}
\caption{Overall comparison of our \method{} and SOTAs.
(Upper) Generated captions by SmallCap, BLIP-2, and our \method{} for a commonsense-violating image from the WHOOPS dataset. \red{\( \times \)} and \blue{\( \checkmark \)} indicate incorrect and correct predictions, respectively. Incorrect objects in captions are highlighted in \colorbox{backred}{red}, while correct ones are in \colorbox{backblue}{blue}. SmallCap and BLIP-2 give incorrect predictions for \textit{``ice skates"} and \textit{``wood floor"}, respectively, while our \method{} utilizes an external visual--name memory to enhance attention to objects within the image, leading to superior performance for image captioning. 
(Lower) Comparison of the number of trainable parameters, CIDEr score on COCO and NoCaps datasets. The size of each circle reflects the log number of trainable parameters. \method{} (3.97M) has less trainable parameters than others while achieving comparable results with SOTAs at scale.  
}
\label{fig:teaser}
\end{figure}

Retrieval-augmented image captioning~\cite{ramos2023smallcap,hu2023reveal} is emerging as an alternative since it considerably reduces training costs in both time and data 
while producing encouraging results.
Nonetheless, with their huge datastore, it is obvious that LLMs would imitate the given texts, limiting their ability to describe open-world objects properly.
For instance, SmallCap~\cite{ramos2023smallcap} considers the words \textit{``skateboard"} and \textit{``wooden floor"} to be a pair regardless of visual appearances containing a commonsense-violating pair of \textit{``ice skates"} and \textit{``wood floor"} (Fig.~\ref{fig:teaser}, upper).
Additionally, prompting the LLMs given a lot of retrieved texts becomes cumbersome, requiring more trainable parameters.
Fig.~\ref{fig:teaser} (lower) shows that 
the CIDEr scores obtained by a lightweight SmallCap~\cite{ramos2023smallcap} with 43M trainable parameters are far away from those obtained by a heavy REVEAL~\cite{hu2023reveal} with 2.1B trainable parameters.
Beyond that, due to the frequent occurrence of new objects, access to their sample texts is not always feasible, making the memory utilized in~\cite{ramos2023smallcap,hu2023reveal} difficult to grow.
We thus aim to streamline the external memory used in previous work~\cite{ramos2023smallcap,hu2023reveal} by storing a sufficiently small amount of object information.
And, of course, not only does the model not stereotype the example sentences, but the number of trainable parameters would be reduced drastically as a result of the causation (Fig.~\ref{fig:teaser}).

We follow~\cite{vo2022noc,fan2023rca} to construct a key-value memory where the key is represented by object's features, and the value corresponds to object's name. Unlike \cite{vo2022noc,fan2023rca}, which 
rely on object definition as the key,
our method leverages the visual appearance of the object
as the key
because of the abundance of object images readily available on the internet. We propose an external visual--name memory tailored for ease of expansion and cost-effectiveness in upholding up-to-date object information.
We present a highly effective retrieval-augmented LLMs-based image captioning method, called \method{}, that prompts frozen LLMs with object names retrieved from our proposed memory for open-world comprehension.
\method{} contains a frozen image encoder ViT~\cite{fang2023eva} and Q-Former~\cite{li2023blip2} with \textit{trainable} image query tokens for object retrieval, an attentive fusion module, a \textit{trainable} linear layer for mapping between vision and language latent spaces, and a frozen LLM decoder~\cite{chiang2023vicuna} for generating captions.
Specifically, the attentive fusion module feeds retrieved object names and visual features into a customized frozen Q-Former using \textit{trainable} object name query tokens to implicitly reduce the presence of superfluous object names.
As a result, \method{} amounts to only 3.97M trainable parameters.
Once trained, the model can be adapted
to new domains and large-scale data without further fine-tuning or re-training.
Our contributions are as follows:

\begin{itemize}[leftmargin=*]

\item 
We provide an extensible external visual--name memory with minimal but useful object information, which enables LLMs-based models to comprehend the open world.

\item We present a highly efficacious retrieval-augmented image captioning \method{} with 3.97M trainable parameters.

\end{itemize}

On in-/out-domain benchmarks and synthetic commonsense-violating dataset, \method{} trained solely on COCO dataset competes with other lightweight methods by a margin while being on par with other specialist SOTAs.

\section{Related Work}

\noindent \textbf{Image captioning} aims to describe the contents of a given image. It can be roughly divided into 
two approaches: non-LLMs-based methods and LLMs-based ones.
The former approaches~\cite{karpathy2015deep,xu2015show,anderson2018bottom} typically employ a visual encoder and a language decoder in an end-to-end fashion to generate captions.
However, they are incapable of describing open-world objects.
The latter one leverages pre-trained large-scale vision models (CLIP~\cite{radford2021clip}, ViT~\cite{dosovitskiy2021vit}) and LLMs (GPTs~\cite{radford2019gpt2,brown2020gpt3}, T5~\cite{raffel2020t5}, LLaMA~\cite{touvron2023llama}) by bridging the gap between two modalities using either pre-training with large-scale data or the learned mapper or prompt techniques.
LLMs-based models~\cite{mokady2021clipcap, li2023blip2, chen2023pali, chen2023palix} demonstrate advancements in image captioning challenges, allowing the capacity to describe anything as long as pre-trained vision models can recognize it.
Our method belongs to the LLMs-based approaches, but instead of relying fully on the pre-trained vision model, we use object names retrieved from the external memory to augment LLMs-based image captioning.

\noindent \textbf{Novel object captioning} is a branch of image captioning that describes images containing objects that were not seen during training.
Non-LLMs-based methods explore more objects by learning from unpaired image-sentence sources (DCC~\cite{hendricks16dcc}, NOC~\cite{venugopalan2017noc}) or rely on novel object detectors to recognize novel concepts (NBT~\cite{lu2018nbt}, OSCAR~\cite{li2020oscar} and VinVL~\cite{zhang2021vinvl}). 
LLMs-based methods such as ViECap~\cite{fei2023viecap} leverage the pre-trained CLIP~\cite{radford2021clip} to obtain object entities.
Nevertheless, the cut-off in training time of the pre-trained object detector or CLIP prevents it from detecting novel objects that arise quickly in reality.
Unlike earlier work, we can readily update our recognition of novel concepts by adding them to external memory, ensuring that we keep any new objects from the past and even the future.

\noindent \textbf{Retrieval-augmented image captioning} is a recently popular approach that augments the captioning model with retrieved information for better open-world understanding. 
AoANet~\cite{fei2021memory} uses 
a memory bank of image-sentence pairs and target words.
SmallCap~\cite{ramos2023smallcap} employs image-to-text retrieval to obtain sampled captions from a captions datastore.
RA-CM3~\cite{yasunaga2023racm3} retrieves documents from an external memory of a mixture of text and image via a dense multimodal retriever. 
EXTRA~\cite{ramos2023retrieval} and Re-ViLM~\cite{yang2023revilm} exploit the similarity of the input image and vision candidates to retrieve captions. 
Different from the previous methods, our external memory contains visual--name pairs to avoid redundant information in the external captions/documents. 
In addition, we use an attentive fusion module to mitigate the effects of irrelevant retrieved object names on caption generation.

\begin{figure*}[ht!]
\centering\includegraphics[width=0.90
\textwidth]{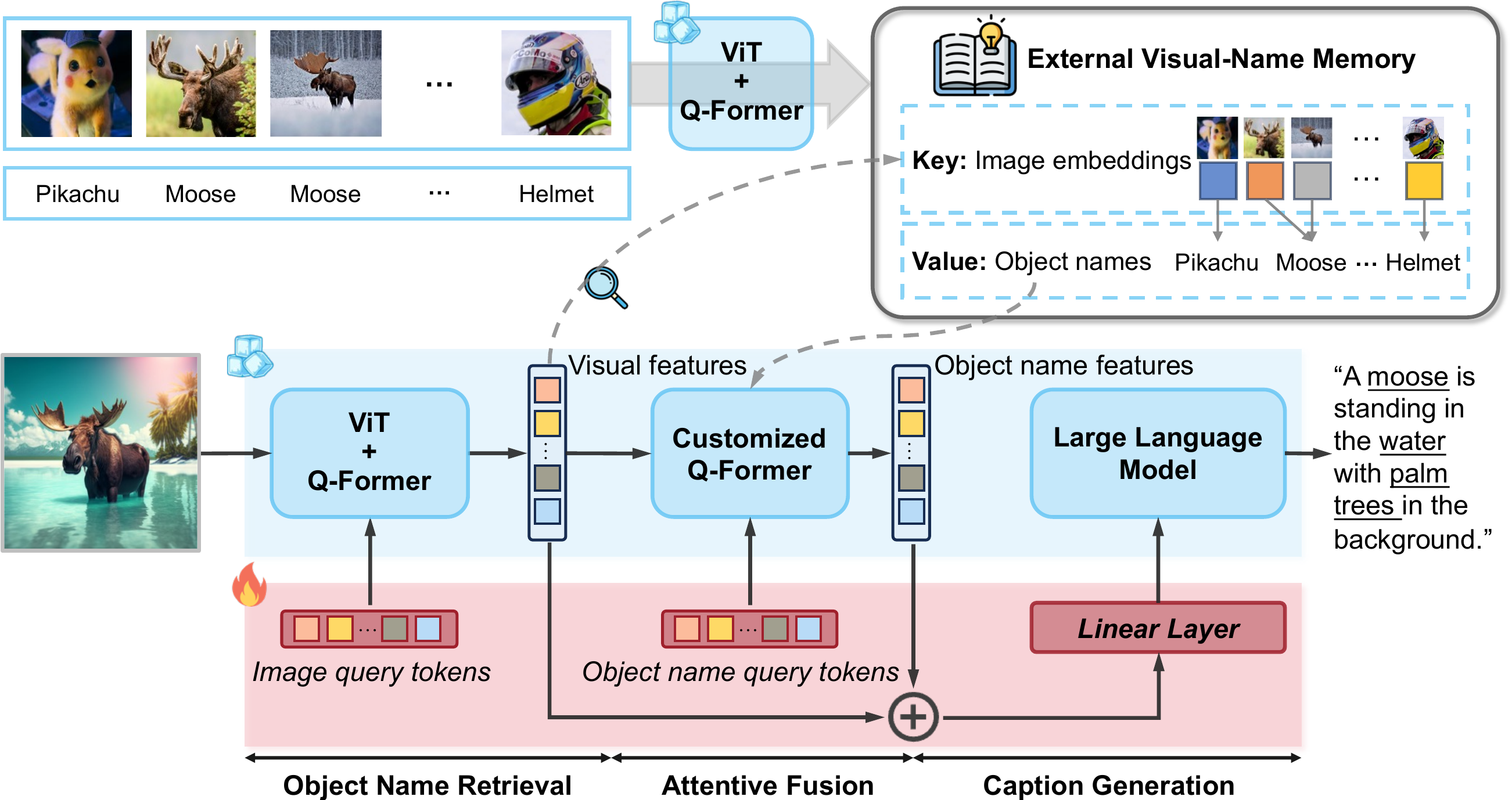}
\caption{Schematic of our proposed \method{}. It consists of an external visual--name memory with image embeddings and object names (upper),  a frozen ViT and Q-Former equipped with \textit{trainable} image query tokens, an attentive fusion module developed by a customized frozen Q-Former and \textit{trainable} object name query tokens, and a frozen LLM with a \textit{trainable} linear layer (lower). The ViT and Q-Former extract learned visual features from the input image, which are then used to retrieve object names from the external memory. These retrieved object names and learned visual features undergo cross-attention in the customized Q-Former, creating refined object name features. Finally, the object name features combined with visual features are fed into the LLM post a linear layer for generating captions.} 
\label{fig_model}
\end{figure*}

\section{Proposed \method{}}

\subsection{Idea of \method{}}
\label{sec:idea}
We aim to build a retrieval-augmented LLMs-based image captioning model with a sufficiently small yet informative external memory.
It involves two challenges: (1) constructing an expandable external memory and (2) building an effective LLMs-based model using retrieved object names.

As discussed above, challenge (1) can be resolved by utilizing the visual appearance of objects. 
However, if we restrict our memory to only a visual--name pair for each object, our memory will be lacking in diversity.
Therefore, we gather several images for each target object.
Additionally, we keep the synthetic images in our memory to avoid the harm that synthetic images might cause to our method, as pointed out in~\cite{hataya2023will}. 
With the capability to collect images from the internet, \method{} can be easily expanded to include novel objects from the real world effortlessly.

We base our method on the frozen pre-trained vision model and LLM with several trainable layers (Fig.~\ref{fig_model}), giving in a model that is cheap to train.
To guide the LLM, we adopt a recently popular approach called prompting as in~\cite{mokady2021clipcap, li2023blip2, dai2023instructblip, ramos2023smallcap, zhu2023minigpt}.
We begin by matching the learned
visual features from the input image with image embeddings stored in memory, retrieving object names.
We also introduce an attentive fusion module designed to implicitly remove irrelevant retrieved names.
Finally, following the attentive fusion, we combine the learned visual features and object name features to form a prompt for the LLM to generate a caption, thus addressing challenge (2).

\subsection{External visual--name memory}
\label{sec_memory}

To build the external visual--name memory, we first collect image--name pairs from the external data source. After that, we encode these images into image embeddings, which serve as keys in memory, and use their names as values.

\noindent \textbf{External data source.}
We utilize object images from LVIS dataset~\cite{gupta2019lvis} to construct our external visual--name memory $\mathcal{M}$.
Specifically, we use 1203 objects in LVIS, where we randomly select from one to ten images for each object, amounting to 8581 object images.
Furthermore, as mentioned in Sec.~\ref{sec:idea}, we also incorporate synthetic images in our memory construction.
Using stable diffusion~\cite{Rombach_2022_diffusion}, we generate five additional images for each object, with a prompt of ``\textit{a photo of \{object name\}}'', resulting in a total of $M=14596$ $(8581 + 5 \times 1203)$ images.
Each object image  $X^i$  is associated with an object name $v^i$.
Note that many object images may share the same object name.
For the sake of simplicity, we may regard each image as corresponding to a single name.
In summary, we have $M$ image--name pairs $\{(X^i,v^i)\}^{M}_{i=1}$ for external memory construction.

\noindent \textbf{External memory construction.} 
For each image $X^i$, we use a frozen vision encoder $\mathcal{E}(\cdot)$ (see Sec.~\ref{sec_retrieval} for detail) to project it into 32 embeddings with the size of $1\times768$ each: $\{\mathbf{k}^i_1,\mathbf{k}^i_2, \cdots, \mathbf{k}^i_{32}\}=\mathcal{E}(X^i)$.
We then average 32 embeddings to produce a single embedding $\mathbf{k}^i$ ($1\times768$) that serves as the key (visual) in $\mathcal{M}$.
The paired object name $v^i$ 
acts as its value (name).
Consequently, we have the visual--name memory $\mathcal{M}=\{(\mathbf{k}^i, v^i)\}^{M}_{i=1}$ which is indexed using FAISS~\cite{johnson2019faiss}, facilitating rapid searches based on similarity measures. 
Our memory can be expanded effortlessly by gathering additional visual--name pairs (see Sec.~\ref{sec:detailed analysis}).

\subsection{Object names retrieval}
\label{sec_retrieval}

\noindent
\textbf{Image encoding.} We feed a frozen vision encoder $\mathcal{E}$ image $X$ and image query tokens $\mathbf{T}_\mathrm{img}$ to produce visual features $\mathcal{Q}$.
To enable the retrieval process controllable, we make image query tokens to be trainable.
Thus, the image encoding process can be summarized as $\mathcal{Q}=\mathcal{E}(X, \mathbf{T}_\mathrm{img})$.
We use the BLIP-2 pre-trained vision encoder~\cite{li2023blip2}, which consists of a pre-trained vision transformer ViT-g~\cite{fang2023eva} outputting image features ($257 \times 1408$), and a Q-Former receiving image features producing $|\mathcal{Q}|=32$ learned visual features ($1 \times 768$ each). We denote $\mathcal{Q}=\{\mathbf{q}_1,\mathbf{q}_2, ..., \mathbf{q}_{32}\}$.

\noindent \textbf{Retrieval.} 
Having obtained $\mathcal{Q}$, we calculate the cosine similarity between the query $\mathbf{q}_{j}\in\mathcal{Q}$ and the key $ \mathbf{k}^i\in\mathcal{M}$. The similarity calculation is given by  $\operatorname{SIM}(\mathbf{q}_{j}, \mathbf{k}^i)=\frac{\mathbf{\mathbf{q}_{j}^{\top}} \mathbf{k}^i}{\|\mathbf{q}_{j}\|\|\mathbf{k}^i\|}$,
where $i\in [1, M]$, $j\in [1, 32]$.
Given each $\mathbf{q}_{j}$, we select one key with the highest similarity score, resulting in 32 key--value candidates $\{\mathbf{k}_j^{\mathrm{best}}, v_j^{\mathrm{best}}\}^{32}_{j=1}$.

After that, we filter out candidates with repeated object names (values), and then select the top-K values.
In particular, we determine the index $j$
from the key that has the highest $\operatorname{SIM}$ score. These selected values $v_j^{\mathrm{best}}$ are redefined as the new notation $v_l$ in the retrieved top-K object names for the input image, which can be summarized as follows:

\begin{equation}
\{\mathbf{k}_j^{\mathrm{best}}, v_j^{\mathrm{best}}\} =\arg \max _{\mathbf{k}^i} \operatorname{SIM}\left(\mathbf{q}_j, \mathbf{k}^i\right), \nonumber\\
\end{equation}

\begin{equation}
j = \arg \max _{j} \operatorname{SIM}(\mathbf{q}_j, \mathbf{k}_j^{\mathrm{best}}),
v_l \gets v_j^{\mathrm{best}}, \nonumber
\end{equation}

\noindent
where $l\in[1, \mathrm{K}]$. As a result, the retrieved top-K object names are $\{v_l\}^\mathrm{K}_{l=1}$.

\subsection{Attentive fusion}
\label{sec_att}

Since the object names obtained from the retrieval process may be redundant, we develop an attentive fusion module to selectively distill object name features.

The retrieved object names $\{v_l\}^\mathrm{K}_{l=1}$ are concatenated together into a sequence $\mathcal{S}$, each separated by a delimiter: 
$\mathcal{S}=\{v_1, \texttt{[SEP]}, v_2, \texttt{[SEP]}, \cdots, \texttt{[SEP]}, v_\mathrm{K} \}$.
The sequence $\mathcal{S}$ and visual features $\mathcal{Q}$ are fed into a customized Q-Former $\mathcal{F(\cdot)}$, which is constructed from the frozen pre-trained Q-Former as we used in vision encoder $\mathcal{E}$.
Nonetheless, in order to enable object names to get attention from visual features, we switch the image embedding port and the text instruction port (see the supplement for architecture detail). 
Like in the image encoding process in Sec.~\ref{sec_retrieval}, 
we make the object name query tokens $\mathbf{T}_\mathrm{obj}$ learnable during training to assist in learning object name features related to the caption. 
The size of $\mathbf{T}_\mathrm{obj}$ is $P \times 768$, where $P$ indicates the number of object name query tokens. 
We get the object name features $\mathcal{V}=\mathcal{F}(\mathcal{S},\mathcal{Q},\mathbf{T}_\mathrm{obj})$. 

\subsection{Caption generation}
\label{sec_llm}

Before inputting the visual features $\mathcal{Q}$ and object name features $\mathcal{V}$ into the LLM decoder, we concatenate ($\oplus$) them and use a linear layer $\phi(\cdot)$ to project them into the input latent space of the LLM as $\phi(\mathcal{Q}\oplus\mathcal{V})$.
The LLM used for caption generation in this work is the pre-trained Vicuna-13B~\cite{chiang2023vicuna}, an open-source chatbot constructed from LLaMA~\cite{touvron2023llama}. During training and evaluation, we design a prompt in a conversational format, that is similar to~\cite{zhu2023minigpt}:

\noindent
\begin{minipage}{0.99\columnwidth}
\centering
\begin{tcolorbox}
\raggedright
\texttt{\#\#\#Human:} \texttt{<Img><ProjFeature></Img>} \texttt{Describe this image in detail.} 
\texttt{\#\#\#Assistant:}
\end{tcolorbox}
\end{minipage}

\noindent
in which, \texttt{ProjFeature} denotes the projected feature $\phi(\mathcal{Q}\oplus\mathcal{V})$ after the linear layer. 
In training phase, given input caption tokens $\{c_i\}^L_{i=1}$, the LLM decoder concatenates the embedded prompt $\{\mathbf{w}_i\}^N_{i=1}$ and 
the embedded caption tokens $\{\mathbf{c}_i\}^L_{i=1}$ as input, and predicts the caption tokens in an autoregressive fashion, while in the evaluation phase, we only need to input the embedded prompt. We train \method{} by minimizing the cross-entropy loss in an end-to-end way: $\mathcal{L}_\theta=-\sum_{i=1}^L \log p_\theta\left(c_i\mid \mathbf{w}_1, ... \mathbf{w}_N, \mathbf{c}_1, ..., \mathbf{c}_{i-1}\right)$, in which $\theta$ indicates the trainable parameters.

\section{Experimental Settings}

\subsection{Training setup}

\noindent \textbf{Implementation.}  \method{} uses the same image encoder as in BLIP-2~\cite{li2023blip2}, consisting of a ViT-g~\cite{fang2023eva} and their pre-trained Q-Former. 
Since we intend to obtain object name features through cross-attention between retrieved object names and visual features, we develop a customized Q-Former, which consists of BERT~\cite{kenton2019bert} with cross-attention layers inserted at every other transformer block.
We use a frozen Vicuna-13B~\cite{chiang2023vicuna} as the caption generator.

\noindent \textbf{Training dataset.} For all experiments, we exclusively train \method{} using the training set of \textbf{COCO} dataset~\cite{lin2014coco}, consisting of 82k images and 5 captions per images. The entire training process takes about 3 hours on 4 A6000 GPUs, using mixed precisions (more details in the supplement).

\input{tables/tab1_overall}

\subsection{Evaluation setup}

\noindent \textbf{Evaluation dataset.} We evaluate \method{}, trained using the COCO training set, across four datasets: its test set, two challenging benchmarks -- NoCaps validation set and Flickr30k test set, and a synthetic commonsense-violating dataset -- WHOOPS.
We adhere follow prior work~\cite{fei2023viecap,wang2023cogvlm} 
to use the same images of Karpathy split~\cite{karpathy2015deep} on \textbf{COCO} test set, \textbf{NoCaps}~\cite{agrawal2019nocaps} validation set, and Karpathy split on \textbf{Flickr30k}~\cite{plummer2015flickr30k} test set.
In addition, \textbf{WHOOPS}~\cite{bitton2023whoops} is a synthetic image captioning dataset comprising 500 synthetic commonsense-violating images and 2500 paired captions.

\noindent \textbf{Compared methods.}
We compare \method{} with several SOTAs. According to the trainable parameters size, they can be divided into 1) Heavyweight-training (between 100M to 5B): VinVL~\cite{zhang2021vinvl}, AoANet~\cite{fei2021memory}, NOC-REK~\cite{vo2022noc}, RCA-NOC~\cite{fan2023rca}, ViECap~\cite{fei2023viecap},
InstructBLIP~\cite{dai2023instructblip}, OSCAR~\cite{li2020oscar}, BLIP~\cite{li2022blip}, BLIP-2~\cite{li2023blip2}, REVEAL~\cite{hu2023reveal}; 
2) Lightweight-training (less than 100M): MiniGPT4~\cite{zhu2023minigpt}, SmallCap~\cite{ramos2023smallcap}, ClipCap~\cite{mokady2021clipcap}; and also 3) Specialist SOTAs with huge trainable parameters (larger than 5B):  Qwen-VL~\cite{bai2023qwen}, CogVLM~\cite{wang2023cogvlm}, PaLI~\cite{chen2023pali}, PaLI-X~\cite{chen2023palix}. Among these methods, AoANet, NOC-REK, RCA-NOC, REVEAL, and SmallCap are retrieval-augmented captioning methods.

\section{Experimental Results}

\subsection{Results on in-/out-domain benchmarks}

\begin{figure*}
\centering
\includegraphics[width=0.88\textwidth]{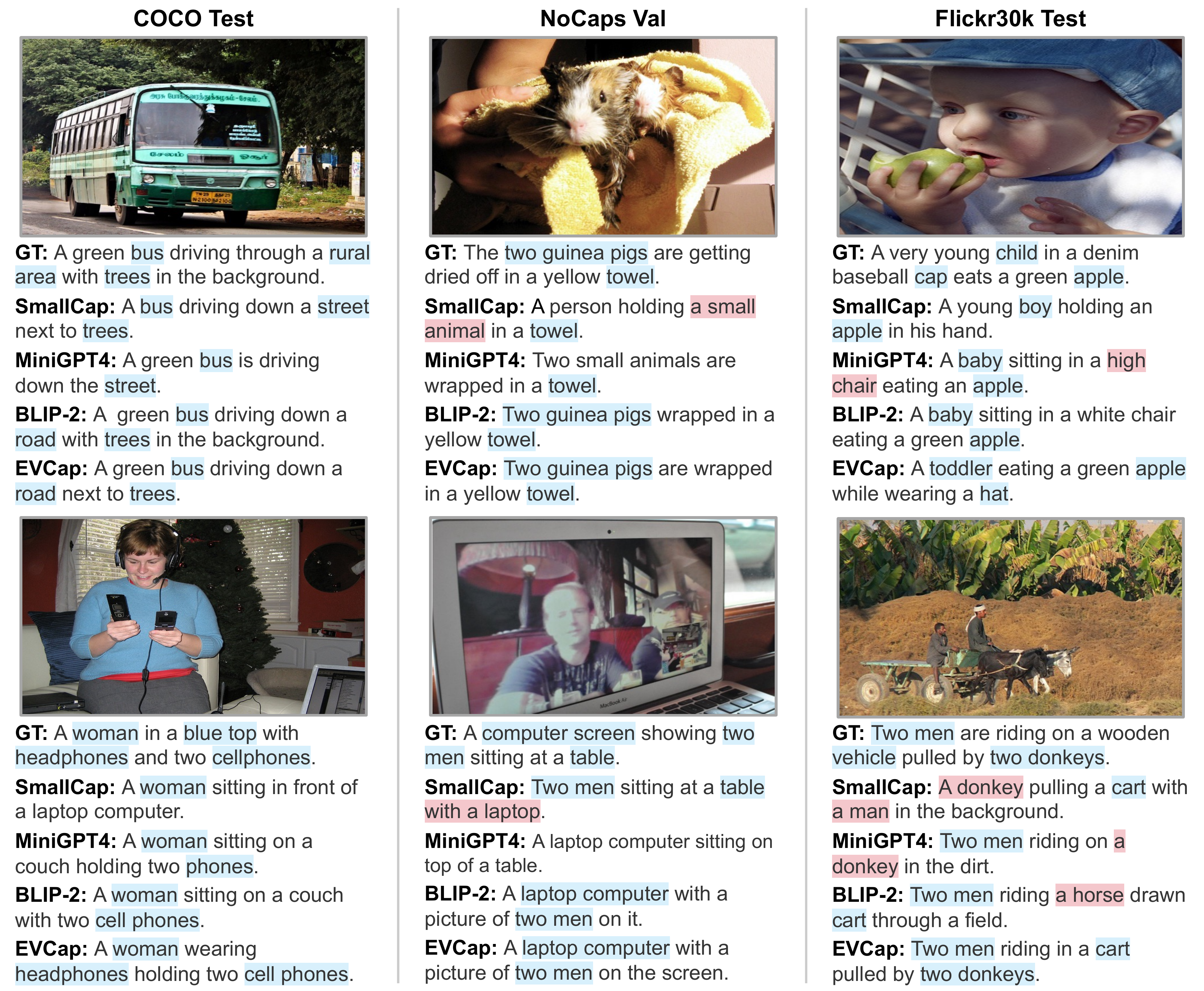}
\caption{Examples of captions generated by our \method{} and three SOTA methods on COCO test set, NoCaps validation set, and Flickr30k test set. GT refers to the Ground Truth captions. Incorrect objects in captions are highlighted in \colorbox{backred}{red}, while correct ones are in \colorbox{backblue}{blue}. Our \method{} correctly generates captions across different datasets, showing performance comparable to BLIP-2.} 
\label{fig:vis_results}
\end{figure*}

We assess \method{} against SOTAs on both in-domain and out-domain benchmarks. The COCO test set can be considered as in-domain data as we only train our model on the COCO training set. Out-domain benchmarks are the NoCaps validation set and the Flickr30k test set.

\noindent \textbf{Quantitative results.}
Tab.~\ref{tab:overall} details our \method{}’s performance in comparison with SOTA methods.
We first evaluate training costs in terms of training data sizes and parameters. 
Similar to various heavyweight-training models that exclude LLMs and the majority of lightweight-training models, \method{} is trained solely on the COCO training set. It utilizes only 3.97M trainable parameters, positioning it as the second smallest, slightly larger than MiniGPT4 with 3.94M.
Among lightweight-training models, our approach outperforms others, achieving the highest scores on all benchmarks. Despite using less training data and nearly identical trainable parameters as MiniGPT4, \method{} significantly surpasses it, with a marked improvement of 10.5, 10.5, and 6.0 in CIDEr scores for each benchmark.
When further compared with heavyweight-training models, the performance of \method{} stands out among million-level models, nearly matching InstructBLIP, except in NoCaps. 
Note that since BLIP-2 does not include Vicuna checkpoints, InstructBLIP performs pre-training with Vicuna using the same procedure as BLIP-2, whereas \method{} does not involve pre-training.
Against REVEAL, which also uses external memory, our \method{} utilizes about 1/3000 training data and 1/500 training parameters yet yields comparable results.
Moreover, \method{}’s performance is on par with BLIP-2, the top-performing model with 1.2B trainable parameters. This highlights \method{}’s efficiency and effectiveness despite its significantly smaller training cost, thanks to our external visual--name memory.
Regarding specialist SOTAs, they use billion-level training data and over 5B trainable parameters, so it is acceptable that they can achieve exceptionally strong performance, surpassing \method{} by nearly 10 on all benchmarks in CIDEr scores.

\noindent \textbf{Qualitative results.} 
Fig.~\ref{fig:vis_results} presents a comparison of captions generated by our \method{} and three SOTA models across three benchmarks. 
The captions of SmallCap are generated by its publicly accessible demo~\cite{smallcapdemo}.
We generate captions of MiniGPT4 and BLIP-2 using their respective pre-trained models. As a lightweight and retrieval-augmented captioning method, SmallCap struggles to produce accurate captions for given images, primarily because it relies on retrieved captions laden with extraneous information. MiniGPT4, though aligned with the primary content of images, sometimes misses certain objects like \textit{``trees"} and \textit{``headphones"}. This oversight stems from its focus on the main objects in images, without integrating additional cues for other objects provided by the retrieved object names. In contrast, the captions generated by our \method{} are comparable to those of BLIP-2.

\subsection{Results on commonsense-violating data}

To explore our \method{}'s capability in describing contents in open-word settings, we further evaluate it on WHOOPS dataset, which contains commonsense-violating images.

\begin{figure}
\centering
\includegraphics[width=0.475\textwidth]{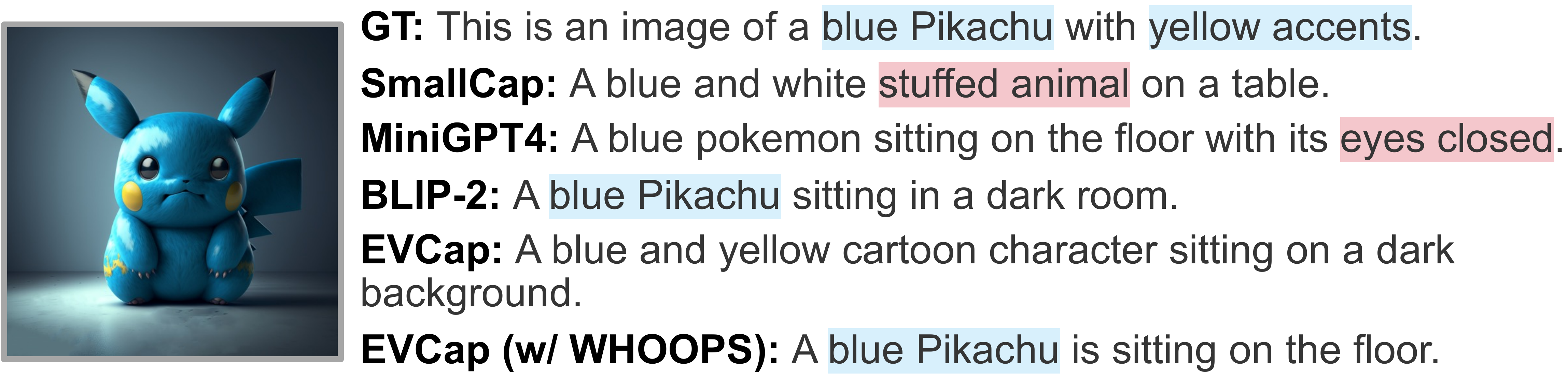}
\caption{Examples of captions generated by our \method{}, \method{} (w/ WHOOPS), and three SOTAs on WHOOPS dataset. Incorrect objects are highlighted in \colorbox{backred}{red}, while correct ones are in \colorbox{backblue}{blue}.}
\label{fig:vis_whoops}
\end{figure}

\input{tables/tab2_whoops}

\noindent \textbf{Quantitative results.} In Tab.~\ref{tab:whoops}, we compare the performance of \method{}, MiniGPT4, BLIP, and BLIP-2 on WHOOPS dataset.
This dataset is particularly challenging due to its inclusion of unusual objects~\cite{bitton2023whoops}. Initially, as an end-to-end trained model, our \method{} exhibits performance similar to MiniGPT4. However, there is a noticeable improvement in the CIDEr score, after the external memory is enriched with 2396 new objects from the WHOOPS dataset, each represented by 5 synthesized images generated using stable diffusion~\cite{Rombach_2022_diffusion}. It highlights the effectiveness of our idea of incorporating an expandable external memory into the captioning model for open-world comprehension.

\noindent \textbf{Qualitative results.}
Fig.~\ref{fig:vis_whoops} illustrates the captions generated by \method{}, \method{} (w/ WHOOPS), and three SOTAs for one image from the WHOOPS dataset.
Similar to other methods except for BLIP-2, \method{} cannot recognize \textit{``blue cartoon character"} as \textit{``Pikachu"}, while \method{} (w/WHOOPS) successfully predicts it because of the updated memory. 
SmallCap and MiniGPT4 tend to generate captions with hallucinatory objects, a result of commonsense-violating contents present in the images.

\subsection{Detailed analysis}
\label{sec:detailed analysis}

\input{tables/tab3_ablation}

\begin{figure}
\centering
\includegraphics[width=0.475\textwidth]{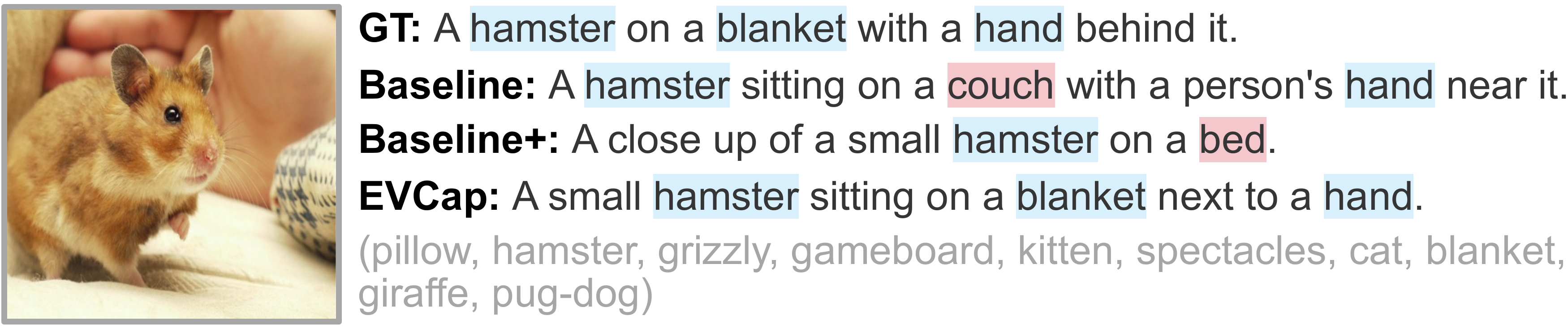}
\caption{Visualization of the captions generated from ablation study on the NoCaps validation set. We also show the retrieved object names by \method{}, presented in \demph{gray}. Incorrect objects in captions are highlighted in \colorbox{backred}{red}, while correct ones are in \colorbox{backblue}{blue}.}
\label{fig:vis_ablation}
\end{figure}

\noindent \textbf{Ablation study.} We assess the contribution of each component prior to the LLM decoder in \method{} by incrementally integrating the image query tokens and the attentive fusion module into our baseline model. The baseline model comprises a ViT+Q-Former, a linear layer, and a LLM decoder. The quantitative results are shown in Tab.~\ref{tab:ablation}. When employing only the baseline model (Baseline), CIDEr scores drop notably by 5.7, 10.5, and 7.6 on COCO, NoCaps, and Flickr30k, respectively. The inclusion of trainable image query tokens (Baseline+) brings a marginal improvement on NoCaps and Flickr30k. However, the performance is significantly enhanced with the addition of attentive fusion (along with the introduction of external memory), indicating the pivotal role of the external visual--name memory in the overall effectiveness of \method{}.
This is further corroborated by the qualitative results in Fig.~\ref{fig:vis_ablation}, where captions from Baseline and Baseline+ inaccurately include objects like \textit{``couch"} and \textit{``bed"}, and Baseline+ overlooks \textit{``hand"}.

\begin{figure}
\centering
\includegraphics[width=0.47\textwidth]{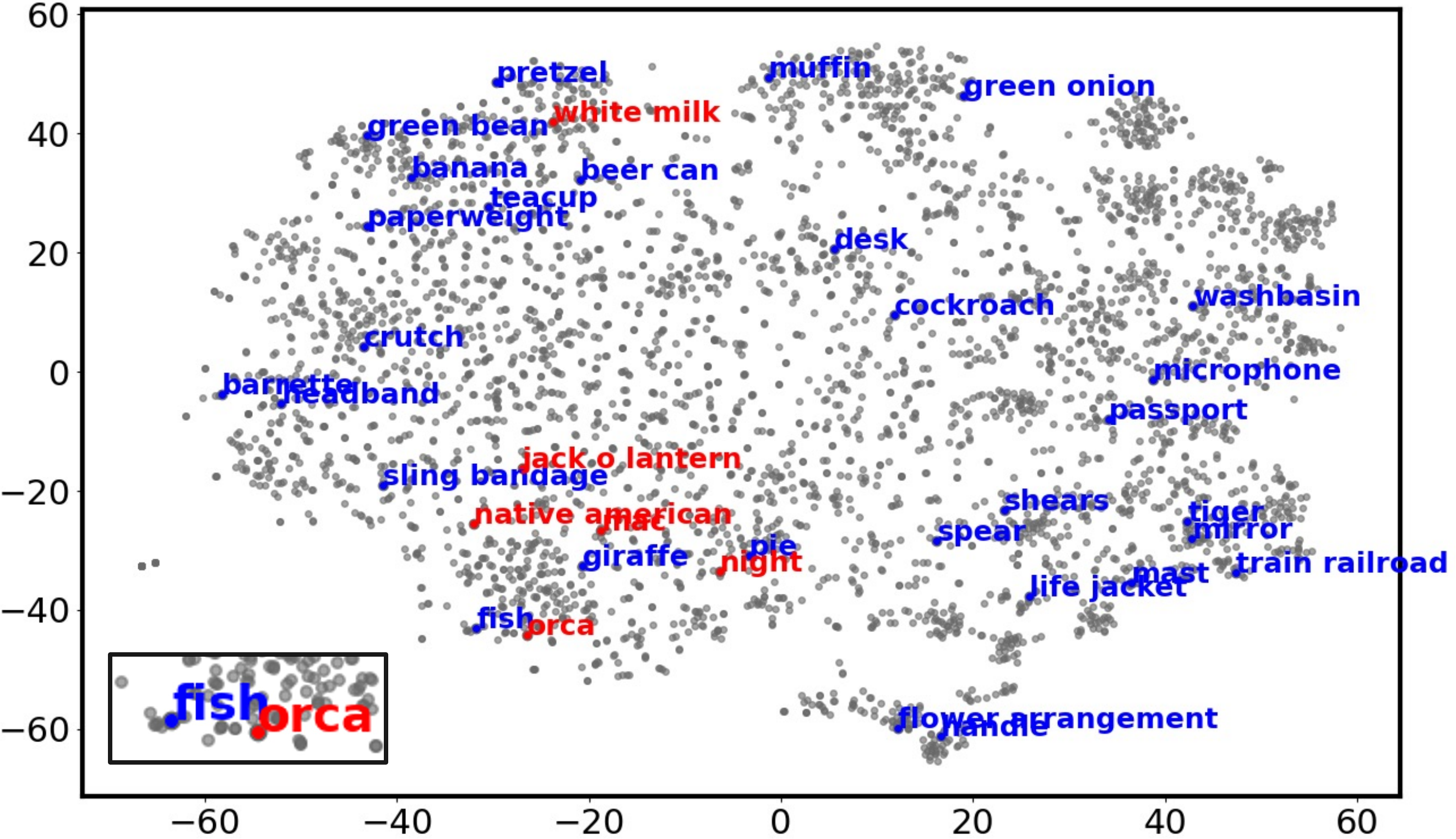}
\caption{Visualization of the visual features in external memory using t-SNE. For visual features in LVIS dataset's objects (\blue{blue}), the related objects fall in the same cluster. After adding more visual features of synthesized images from WHOOPS' objects, new objects (\red{red}) are located at appropriate clusters (zoom-in view).}
\label{fig:vis_tsne}
\end{figure}

\noindent \textbf{Exploration for external memory expandability.} 
To demonstrate the scalability of the external memory in \method{}, we visualize the visual features stored in LVIS external memory, and newly synthesized data from objects appearing in the WHOOPS dataset. We employ t-SNE~\cite{van2008tsne} to plot visual features after reducing their dimensions to 2-D (Fig.~\ref{fig:vis_tsne}). For clear visualization, we only randomly display 3649 visual features in LVIS memory, and add 479 visual features from WHOOPS objects. Among them, 35 samples are randomly labeled. The result shows a clear clustering of LVIS objects (blue) in the external memory, as well as the successful integration and appropriate localization of new objects from WHOOPS (red) into these clusters.
This pattern not only confirms the distinctiveness of visual features already present in the memory but also demonstrates the potential to accurately incorporate and differentiate new objects introduced from updated data. These findings highlight our external memory's ability to expand and maintain its effectiveness even as new data is incorporated.

\noindent \textbf{Impact of external memory size.}
We examine the impact of external memory size in Tab.~\ref{tab:ext_data}. On the one hand, we randomly remove 30\%, 60\%, and 90\% data in the external memory constructed from LVIS objects. The results show the performance gradually degrades on NoCaps as reducing 30\% and 90\% LVIS.
Despite some unexpected increases in certain results on NoCaps (5th row) and Flickr30k (4th - 5th rows), they do not alter the overall downward trend. Similar phenomena are also noted in SmallCap~\cite{ramos2023smallcap}, we speculate it is due to data distribution.
On the other hand, as we infuse WHOOPS knowledge into LVIS memory, there is a slight improvement on NoCaps (out) and Flickr30k. 
These observations validate the model's capability to effectively retrieve object names from an updated memory, enhancing its performance in generating captions.

\input{tables/tab5_external_memory_size}

\begin{figure}[tb]
\centering\includegraphics[width=0.47
\textwidth]{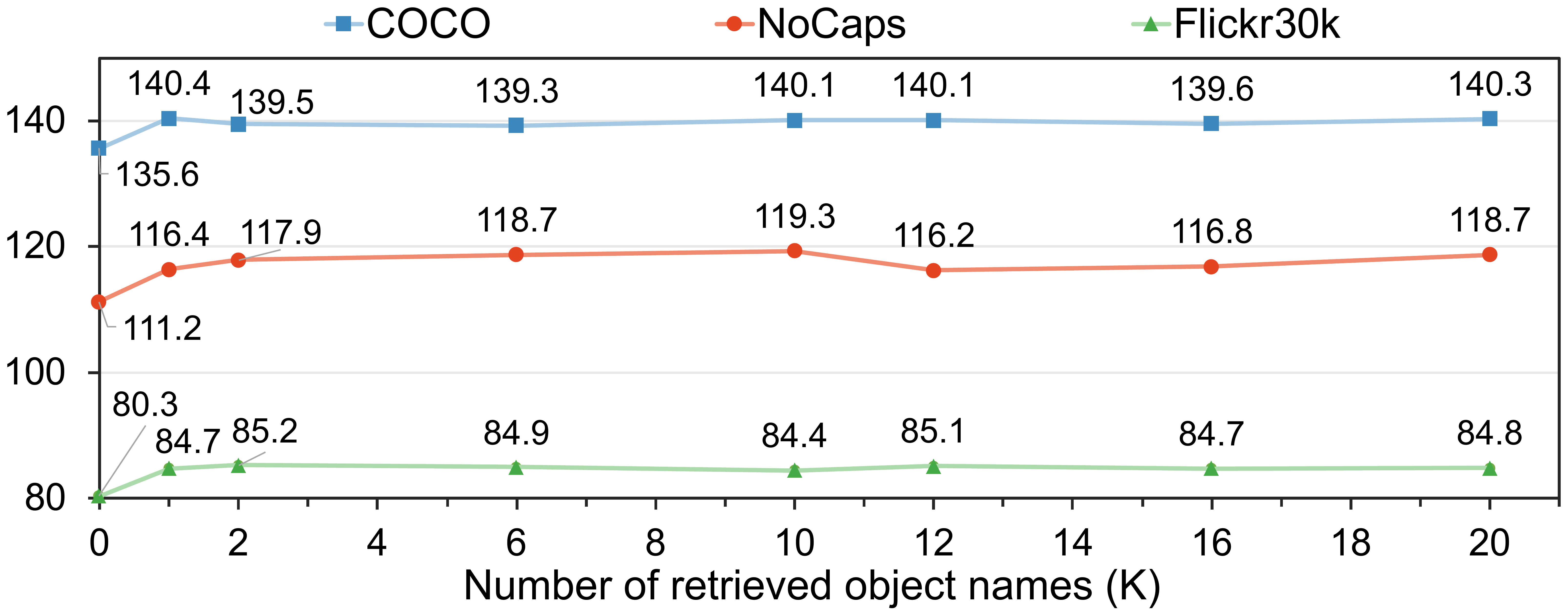}
\caption{CIDEr scores after training \method{} with the number of retrieved object names $\mathrm{K}$ from 0 to 20. The results indicate that the performance is relatively optimal when $\mathrm{K}$ is set to be 10.}
\label{fig:numbers}
\end{figure}

\noindent
\textbf{Impact of the number of retrieved object names.} We investigate how the number of retrieved object names $\mathrm{K}$ (Sec.~\ref{sec_retrieval}) affect \method{} in Fig.~\ref{fig:numbers}. 
We train the model with $\mathrm{K}$ from 0 to 20 and evaluate the performance under CIDEr on all three benchmarks. From the results, we can find that the model works worst on the out-domain dataset (NoCaps) with zero object names. It confirms that the performance boost from Baseline+ to \method{} (Tab.~\ref{tab:ablation}) is primarily attributed to the retrieval-augmented mechanism, but not the customized Q-Former itself. 
With more object names, performance fluctuates but improves.
Furthermore, we observe that setting $\mathrm{K}$ to 10 yields relatively optimal overall performance, validating the choice of $\mathrm{K}=10$ in \method{}.

\noindent \textbf{Analysis with different decoders.} To explore the influence of different LLMs decoders on our \method{}, we experiment by substituting Vicuna-13B with GPT2 and Vicuna-7B, as detailed in Tab.~\ref{tab:llms}. With GPT2 as the decoder, \method{} still markedly surpasses other GPT2-based models, achieving impressive gains of 11.3 and 10.0 under CIDEr on COCO and Flickr30k, compared to SmallCap. When employing Vicuna-7B, the comparison of performance trends mirrors those observed with Vicuna-13B, further attesting to the robustness and adaptability of \method{} across different LLM decoders. Notably, both SmallCap, which retrieves captions, and our GPT2-based \method{}, which retrieves object names, use the same GPT2 decoder. Therefore, their comparison also underscores the effectiveness of our method's object name retrieval and attentive fusion strategy.

\input{tables/tab4_llms}

\noindent \textbf{Limitations.} 

First, \method{} cannot retrieve all objects that appear in the given image due to the memory coverage limits, leading to incomplete image descriptions (Fig.~\ref{fig:vis_whoops}). We will investigate integrating object detection with image captioning to enhance completeness.
Second, our focus on object representation restricts consideration of other crucial captioning elements, affecting overall performance. Similar to all models trained with COCO dataset, \method{} has limitations in generating varied styles, which is reflected in our relatively modest performance improvements in Tab.~\ref{tab:whoops}, compared to MiniGPT4. We will overcome it by exploring methodologies that encourage style diversity in the future.

\section{Conclusion}

We further advance image captioning in real-world scenarios by introducing \method{}, a novel image captioning model with object names retrieved from an external visual--name memory. The external memory is easily expandable, allowing for effortless updates with new object visuals and names. 
We extensively compare \method{} with SOTAs on various benchmarks and commonsense-violating data, demonstrating its significant superiority in performance.

\small{
\noindent
\textbf{Acknowledgements.}
This work was supported by JSPS/MEXT KAKENHI Grant Numbers JP22H05015, JP23H03449, JP23KJ0404, and JP22K17947, and the commissioned research (No.~225) by the National Institute of Information and Communications Technology (NICT), Japan, and the Institute of AI and Beyond of the University of Tokyo, and ROIS NII Open Collaborative Research 2023-23FC01, 2024-24S1201.}

{
    \small
    \bibliographystyle{ieeenat_fullname}
    \bibliography{main}
}

\newpage
\setcounter{section}{0}
\setcounter{figure}{0}
\setcounter{table}{0}

\input{supp.tex}

\end{document}

%% file: preamble.tex
\usepackage[dvipsnames]{xcolor}
\newcommand{\red}[1]{\textbf{\color{red}#1}}
\newcommand{\blue}[1]{\textbf{\color{blue}#1}}

%% file: tables/tab1_overall.tex
\begin{table*}[tb]
\centering
\caption{Quantitative comparison against SOTA methods on three common image captioning benchmarks. * denotes using a \textbf{memory bank}. We report the size of training data and parameters;  BLEU@4 (B@4), METEOR (M), CIDEr (C), and SPICE (S) scores on COCO test set; C and S scores on in-domain, near-domain, out-domain and overall data of NoCaps validation set; C and S scores on Flickr30k test set. Higher score is better. \colorbox{backblue1}{\textbf{Bold}} indicates the best results among compared methods, \colorbox{backblue}{normal} indicates the second best results.} 
\vspace*{-0.5\baselineskip}	
\label{tab:overall}
\resizebox{\linewidth}{!}{
\begin{tabular}{l|cc|cccc|cccccccc|cc}
\toprule[1pt]
\multirow{3}{*}{\textbf{Method}} & \multicolumn{2}{c|}{\textbf{Training}} & \multicolumn{4}{c|}{\textbf{COCO}} & \multicolumn{8}{c|}{\textbf{NoCaps val}} & \multicolumn{2}{c}{\textbf{Flickr30k}} \\
& \multicolumn{1}{c|}{Data} & \multicolumn{1}{c|}{Para.}
& \multicolumn{4}{c|}{Test}
& \multicolumn{2}{c|}{In-domain} & \multicolumn{2}{c|}{Near-domain} 
& \multicolumn{2}{c|}{Out-domain} & \multicolumn{2}{c|}{Overall}
& \multicolumn{2}{c}{Test} \\
& & 
& \multicolumn{1}{c}{B@4} & \multicolumn{1}{c}{M} & \multicolumn{1}{c}{C} & \multicolumn{1}{c|}{S}
& \multicolumn{1}{c}{C} & \multicolumn{1}{c|}{S} 
& \multicolumn{1}{c}{C} & \multicolumn{1}{c|}{S} 
& \multicolumn{1}{c}{C} & \multicolumn{1}{c|}{S} 
& \multicolumn{1}{c}{C} & \multicolumn{1}{c|}{S} 
& \multicolumn{1}{c}{C} & \multicolumn{1}{c}{S} \\
\midrule
\multicolumn{17}{l}{\textbf{Heavyweight-training models}} \\
VinVL~\cite{zhang2021vinvl} & 8.9M  & 110M  &38.2 &30.3 &129.3 &\colorbox{backblue}{23.6}  & 96.8 & 13.5 & 90.7 & 13.1 & 87.4 & 11.6 & 90.9 & 12.8 & -- & --  \\
AoANet+MA*~\cite{fei2021memory} & COCO  & --  & 38.0 & 28.7 & 121.0  & 21.8 & --  & -- & --  & --  & -- & -- & --  & --  & -- & --   \\
NOC-REK*~\cite{vo2022noc} & COCO & 110M & -- & -- & --  & --  & 104.7 & 14.8 & 100.2 & 14.1 & 100.7 & 13.0 & 100.9 & 14.0  & -- & --  \\
RCA-NOC*~\cite{fan2023rca}  & COCO & 110M & 37.4 & 29.6 & 128.4  & 23.1  & 92.2 & 12.9 & 87.8 & 12.6 & 87.5 & 11.5 & 88.3 & 12.4 & -- & -- \\
ViECap \demph{$_\text{GPT2}$}~\cite{fei2023viecap} & COCO   & 124M & 27.2 & 24.8 & 92.9  & 18.2  & 61.1 & 10.4 & 64.3 & 9.9 & 65.0 & 8.6 & 66.2 & 9.5  & 47.9 & 13.6 \\
InstructBLIP \demph{$_\text{Vicuna-13B}$}~\cite{dai2023instructblip} & 129M  & 188M & -- & -- & --  & --  & --  & --  & --  & -- & --  & --  & \colorbox{backblue}{121.9}  & --  & \colorbox{backblue}{82.8} & -- \\
OSCAR~\cite{li2020oscar} & 4.1M  & 338M  & 37.4 & \colorbox{backblue}{30.7} & 127.8 & 23.5 & 83.4 & 12.0 & 81.6 & 12.0 & 77.6 & 10.6 & 81.1 & 11.7 & -- & -- \\
BLIP~\cite{li2022blip}  & 129M  & 446M & 40.4 & -- & 136.7 & -- & \colorbox{backblue}{114.9} & 15.2 & 112.1 & 14.9 & 115.3 & 14.4 & 113.2 & 14.8 & -- & -- \\
BLIP-2 \demph{$_\text{FlanT5-XL}$}~\cite{li2023blip2}  & 129M  & 1.2B & \colorbox{backblue1}{\textbf{42.4}} & -- & \colorbox{backblue}{144.5} & -- & \colorbox{backblue1}{\textbf{123.7}} & \colorbox{backblue1}{\textbf{16.3}} & \colorbox{backblue1}{\textbf{120.2}} & \colorbox{backblue1}{\textbf{15.9}} & \colorbox{backblue1}{\textbf{124.8}} & \colorbox{backblue1}{\textbf{15.1}} & 121.6 & \colorbox{backblue1}{\textbf{15.8}}  & -- & -- \\
REVEAL* \demph{$_\text{T5}$}~\cite{hu2023reveal}  & 1.3B  & 2.1B  & -- & -- & \colorbox{backblue1}{\textbf{145.4}} & --  & --  & --  & --  & -- & --  & --  & \colorbox{backblue1}{\textbf{123.0}} & --  & -- & -- \\
\midrule
\multicolumn{17}{l}{\textbf{Lightweight-training models}} \\
MiniGPT4 \demph{$_\text{Vicuna-13B}$}~\cite{zhu2023minigpt} & 5M  & \colorbox{backblue1}{\textbf{3.94M}} & 38.0 & 29.6 & 129.6  & 23.4  & 99.0  & 14.8  & 106.9  & 15.3 & 110.8  & \colorbox{backblue}{14.9}  & 108.8 & 15.1  & 78.4 & \colorbox{backblue}{16.9} \\
SmallCap* \demph{$_\text{GPT2}$}~\cite{ramos2023smallcap} & COCO  & 7M & 37.0 & 27.9 & 119.7  & 21.3  & --  & --  & --  & -- & --  & --  & -- & --  & 60.6 & -- \\
ClipCap \demph{$_\text{GPT2}$}~\cite{mokady2021clipcap}  & COCO  & 43M & 33.5 & 27.5 & 113.1  & 21.1   & 84.9 & 12.1 & 66.8 & 10.9 & 49.1 & 9.6 & 65.8 & 10.9    & -- & -- \\
\rowcolor{Gray}
\method* \demph{$_\text{Vicuna-13B}$} & COCO  & \colorbox{backblue}{3.97M} & \colorbox{backblue}{41.5} & \colorbox{backblue1}{\textbf{31.2}} & 140.1  & \colorbox{backblue1}{\textbf{24.7}}  & 111.7 & \colorbox{backblue}{15.3} & \colorbox{backblue}{119.5} & \colorbox{backblue}{15.6} & \colorbox{backblue}{116.5} & 14.7 & 119.3 & \colorbox{backblue}{15.3} & \colorbox{backblue1}{\textbf{84.4}} & \colorbox{backblue1}{\textbf{18.0}}  \\
\midrule
\multicolumn{17}{l}{\demph{\textbf{Specialist SOTAs}}} \\
\demph{Qwen-VL \demph{$_\text{Qwen-7B}$}~\cite{bai2023qwen}} & \demph{1.4B} & \demph{9.6B} & \demph{--} & \demph{--} & \demph{--}  & \demph{--}  & \demph{--}  & \demph{--}  & \demph{--}  & \demph{--} & \demph{--}  & \demph{--}  & \demph{121.4} & \demph{--} & \demph{85.8} & \demph{--} \\
\demph{CogVLM \demph{$_\text{Vicuna-7B}$}~\cite{wang2023cogvlm}} & \demph{1.5B}  & \demph{6.5B} & \demph{--} & \demph{--} & \demph{148.7}  & \demph{--}  & \demph{--}  & \demph{--}  & \demph{--}  & \demph{--} & \demph{132.6}  & \demph{--}  & \demph{128.3} & \demph{--} & \demph{94.9} & \demph{--} \\
\demph{PaLI \demph{$_\text{mT5-XXL}$}~\cite{chen2023pali}} & \demph{1.6B}  & \demph{17B} & \demph{--} & \demph{--} & \demph{149.1}  & \demph{--}  & \demph{--}  & \demph{--}  & \demph{--}  & \demph{--} & \demph{--}  & \demph{--}  & \demph{127.0}  & \demph{--} & \demph{--} & \demph{--}\\
\demph{PaLI-X \demph{$_\text{UL2-32B}$}~\cite{chen2023palix}} & \demph{2.2B}  & \demph{55B} & \demph{--} & \demph{--} & \demph{149.2}  & \demph{--}  & \demph{--}  & \demph{--}  & \demph{--}  & \demph{--} & \demph{--}  & \demph{--}  & \demph{126.3}  & \demph{--} & \demph{--} & \demph{--} \\
\bottomrule[1pt]
\end{tabular}
}
\end{table*}

%% file: tables/tab2_whoops.tex
\begin{table}[tb]
{\centering
\caption{Quantitative results on commonsense-violating data -- WHOOPS dataset. \method{} (w/ WHOOPS) denotes \method{} using the memory expanded by WHOOPS objects. The results reveal the open-world comprehension ability and expandability of \method.}
\label{tab:whoops}
\vspace*{-0.5\baselineskip}	
\resizebox{\linewidth}{!}{
\begin{tabular}{lcccc}
\toprule[1pt]
\multirow{1}{*}{\textbf{Method}} 
& B@4 & M & C & S\\ 
\midrule
\multicolumn{5}{l}{\textbf{Only pre-trained models}} \\
BLIP ~\cite{li2022blip} (from~\cite{bitton2023whoops}) & 13  & -- & 65 & -- \\ 
BLIP-2 \demph{$_\text{FlanT5-XXL}$}~\cite{li2023blip2} (from~\cite{bitton2023whoops}) & 31  & -- & 120 & -- \\ 
BLIP-2 \demph{$_\text{FlanT5-XXL}$}~\cite{li2023blip2} (reproduced)  & 28  & 26.7 & 93.1 & 17.9 \\ 
\midrule
\multicolumn{5}{l}{\textbf{Finetuned models on COCO}} \\
MiniGPT4~\cite{zhu2023minigpt}  & 24.2   & 26.7  & 84.8 & 18.2 \\ 
BLIP ~\cite{li2022blip} & 22.9  & 25.0 & 79.3 & 17.1 \\ 
BLIP-2 \demph{$_\text{FlanT5-XL}$}~\cite{li2023blip2} & 25.8  & 27.0 & 89.1 & 18.3 \\ 
\midrule
\multicolumn{5}{l}{\textbf{End-to-end trained models on COCO}} \\
\rowcolor{Gray}
\method{} & 24.1   & 26.1 & 85.3 & 17.7 \\ 
\rowcolor{Gray}
\method{} (w/ WHOOPS) & 24.4 & 26.1 & 86.3 & 17.8 \\ 
\bottomrule[1pt]
\end{tabular}
}
}
\end{table}

%% file: tables/tab3_ablation.tex
\begin{table}[tb]
{\centering
\caption{Ablation study on components prior to the LLM decoder in \method{}. The result of ``+ Attentive fusion" demonstrates the substantial impact of the external visual--name memory.}
\label{tab:ablation}
\vspace*{-0.5\baselineskip}	
\resizebox{\linewidth}{!}{
\begin{tabular}{lcccccc}
\toprule[1pt]
\multirow{2}{*}{\textbf{Method}} & 
\multicolumn{2}{c}{\textbf{COCO test}}  & \multicolumn{2}{c}{\textbf{NoCaps val}}  & \multicolumn{2}{c}{\textbf{Flickr30k test}} \\ 
\cmidrule(lr){2-3} \cmidrule(lr){4-5} \cmidrule(lr){6-7} & C & S  & C & S & C & S  \\ 
\midrule
ViT + Q-Former (Baseline) & 134.4   &23.9  & 108.8   & 14.2 & 76.8  & 17.3  \\ 
\hspace{1em} + Image query tokens  (Baseline+) & 134.1   & 23.8  & 109.0   & 14.3 & 77.3   & 17.2 \\ 
\hspace{1em} + Attentive fusion (\method{}) & 140.1   & 24.7  & 119.3   & 15.3  & 84.4   & 18.0 \\ 

\bottomrule[1pt]
\end{tabular}
}
}
\end{table}

%% file: tables/tab5_external_memory_size.tex
\begin{table}[tb]
{\centering
\caption{Impact of the external memory size on the performance of \method{} by evaluation under CIDEr scores. Changes in the size of external memory result in changes in performance.}
\label{tab:ext_data}
\vspace*{-0.5\baselineskip}	
\resizebox{\linewidth}{!}{
\begin{tabular}{lccccc}
\toprule[1pt]
\multirow{2}{*}{\textbf{Method}} & 
\multicolumn{4}{c}{\textbf{NoCaps val}}  &
\multicolumn{1}{c}{\textbf{Flickr30k}} \\ 
\cmidrule(lr){2-5} \cmidrule(lr){6-6} & In & Near  & Out & Overall & Test \\ 
\midrule
LVIS objects (\method{}) & 111.7   & 119.5  & 116.5   & 119.3  & 84.4\\ 
\hspace{1em} -- 30\% LVIS  & 112.0   & 119.2 & 115.3   & 118.8  & 85.0   \\ 
\hspace{1em} -- 60\% LVIS  & 111.4   & 119.1 & 116.2   & 119.0  &   85.1 \\ 
\hspace{1em} -- 90\% LVIS  & 110.6   & 118.2  & 115.8   & 118.3 & 83.6  \\ 
\hspace{1em} + WHOOPS & 110.7   & 118.9 & 116.7  & 119.0 & 84.9  \\ 

\bottomrule[1pt]
\end{tabular}
}
}
\end{table}

%% file: tables/tab4_llms.tex
\begin{table}[tb]
{\centering
\caption{Analysis with different LLM decoders including GPT2, Vicuna-7B, and Vicuna-13B. The results reveal \method{} is effective when applying it in different LLM decoders.}
\label{tab:llms}
\vspace*{-0.5\baselineskip}	
\resizebox{\linewidth}{!}{
\begin{tabular}{llcccccc}
\toprule[1pt]
\multirow{2}{*}{\textbf{Method}} & \multirow{2}{*}{\textbf{LLM}} & 
\multicolumn{2}{c}{\textbf{COCO test}}  & \multicolumn{2}{c}{\textbf{NoCaps val}}  & \multicolumn{2}{c}{\textbf{Flickr30k test}} \\ 
\cmidrule(lr){3-4} \cmidrule(lr){5-6} \cmidrule(lr){7-8} & & C & S  & C & S & C & S  \\ 
\midrule

SmallCap~\cite{ramos2023smallcap} & GPT2 & 119.7  & 21.3  & --  & --  & 60.6 & -- \\
ViECap~\cite{fei2023viecap}& GPT2 & 92.9  & 18.2  & 66.2 & 9.5  & 47.9 & 13.6 \\
\method{} & GPT2 & 131.0   & 23.2  & 97.6   & 13.3 & 70.6   & 16.1  \\ 
\midrule
MiniGPT4~\cite{zhu2023minigpt} & Vicuna-7B & 119.4  & 23.5  & 108.7  & 15.7  & 73.9 & 17.2 \\
InstructBLIP~\cite{dai2023instructblip} & Vicuna-7B & --  & --  & 123.1  & --  & 82.4 & -- \\
\method{} & Vicuna-7B & 139.0   & 24.7  & 116.8   & 15.3  & 82.7   & 18.0 \\ 
\midrule
MiniGPT4~\cite{zhu2023minigpt} & Vicuna-13B & 129.6  & 23.4  & 108.8  & 15.1  & 78.4 & 16.9 \\
InstructBLIP~\cite{dai2023instructblip} & Vicuna-13B & --  & --  & 121.9  & --  & 82.8 & -- \\
\method{} & Vicuna-13B & 140.1   & 24.7  & 119.3   & 15.3  & 84.4   & 18.0\\ 
\bottomrule[1pt]
\end{tabular}
}
}
\end{table}

%% file: supp.tex
\renewcommand\thesection{\Alph{section}}
\renewcommand\thefigure{\Alph{figure}} 
\renewcommand\thetable{\Alph{table}}





%
\definecolor{cvprblue}{rgb}{0.21,0.49,0.74}


\def\paperID{6846} 
\def\confName{CVPR}
\def\confYear{2024}



\twocolumn[
\centering
\section*{
\Large
Supplementary Material for \\\method: Retrieval-Augmented Image Captioning with External Visual--Name Memory for Open-World Comprehension}
\vspace{0.5cm}
]


This supplementary material complements our paper with the following sections: First, we delve into the implementation specifics of our \method{}, which were not covered in the main paper (see Sec.~\ref{sup_sec_imp}). Second, we offer an expanded discussion on the external visual-name memory, as utilized in the main paper (see Sec.~\ref{sup_sec_mem}). Finally, we present additional results to evaluate the effectiveness of \method{} (see Sec.~\ref{sup_sec_anal}).

\section{Implementation Details}
\label{sup_sec_imp}

\begin{figure}[ht!]
\centering
\includegraphics[width=0.47\textwidth]{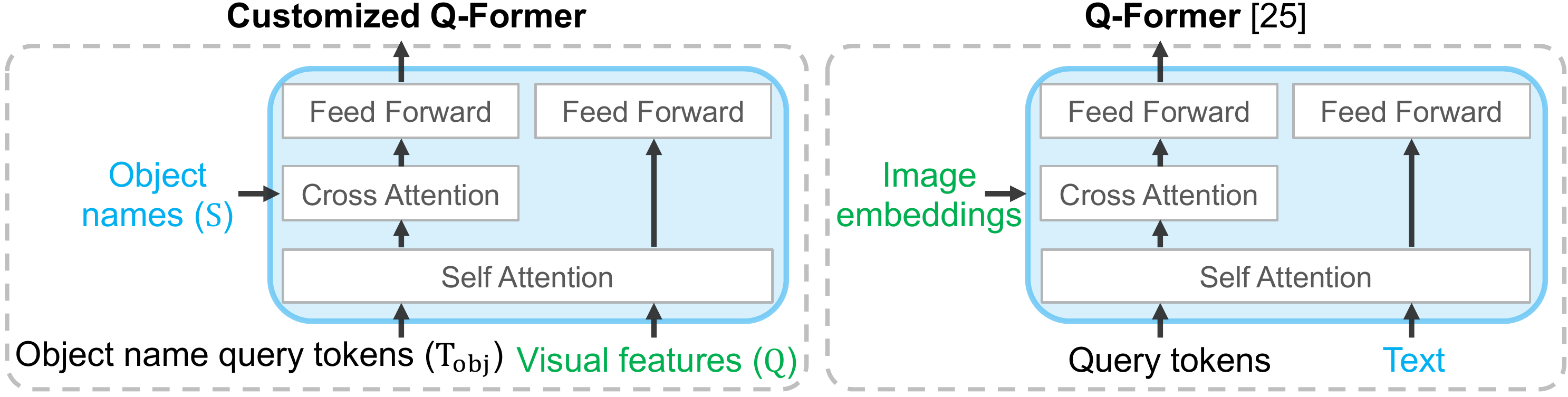}
\vspace*{-0.5\baselineskip}
\caption{Customized Q-Former and original Q-Former~[\textcolor{cvprblue}{25}].} 
\label{fig:qformer}
\end{figure}

\noindent \textbf{Customized Q-Former.}
Fig.~\ref{fig:qformer} depicts customized Q-Former, where the image embedding port (CLIP visual feature port) now receives retrieved object names $\mathcal{S}$, while the text port receives visual features $\mathcal{Q}$.

\noindent \textbf{Implementation.}
Our method is based on Pytorch and is trained within one epoch with a batch size of 24 using mixed precisions.
We optimize the model using AdamW, setting the weight decay at 0.05, and using $\beta_1$ and $\beta_2$ values of 0.9 and 0.99, respectively.
A cosine learning rate (LR) decay strategy is adopted, starting with an initial LR of 1e-4. The model undergoes 5000 linear warm-up steps, beginning with a start LR of 1e-6. During the evaluation phase, we use a beam search strategy with a beam size of 5 to generate captions.

\begin{figure*}[tb]
\centering\includegraphics[width=0.99\textwidth]{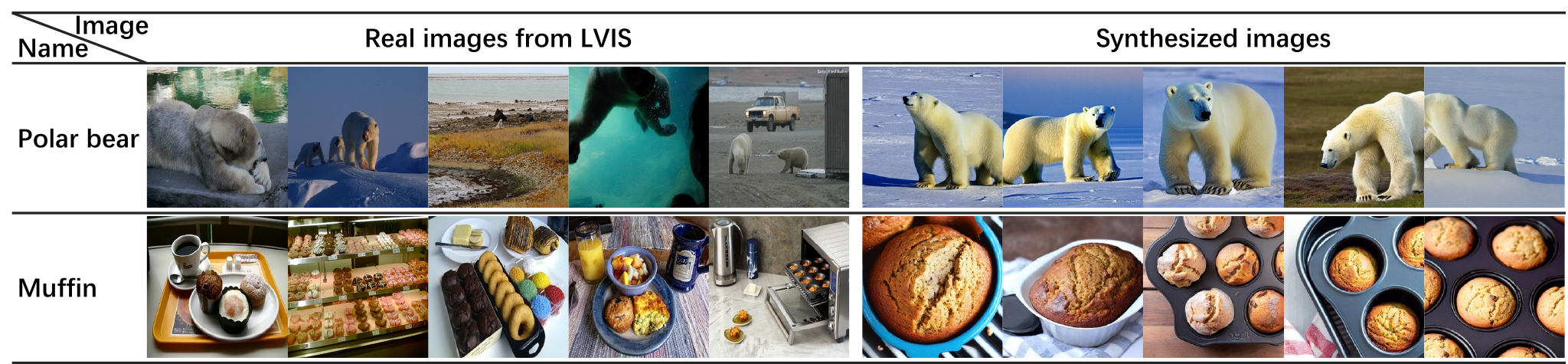}
\caption{Samples of images and corresponding names in the external visual--name memory constructed from LVIS's objects.}
\label{sup_fig_lvis}
\end{figure*}

\begin{figure}[ht!]
\centering\includegraphics[width=0.46\textwidth]{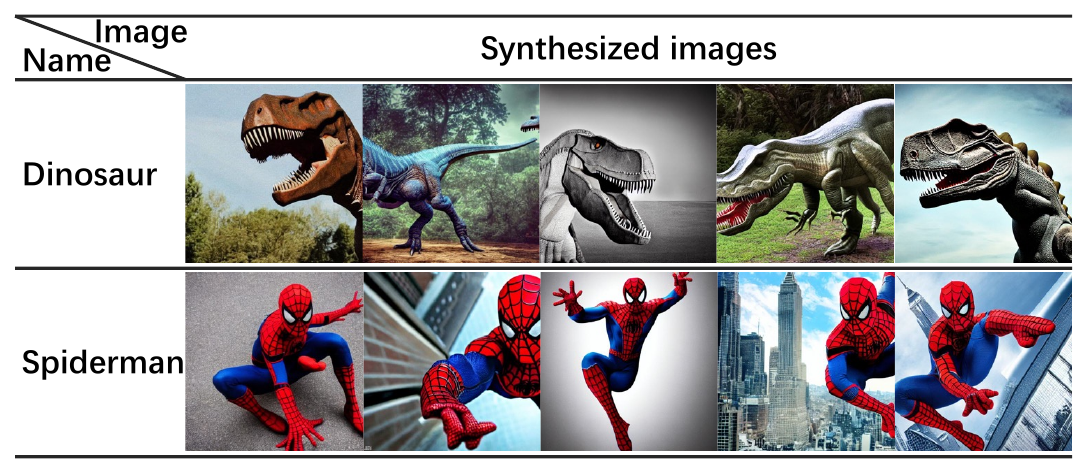}
\caption{Samples of images and corresponding names in the added WHOOPS's objects.}
\label{sup_fig_whoops}
\end{figure}

\section{External Visual--Name Memory}
\label{sup_sec_mem}

\subsection{LVIS memory}
As stated in Sec.~\textcolor{red}{3.2} of the main paper, we utilize 1203 objects from the LVIS dataset. For each of these objects, we randomly select between one and ten images from LVIS. Additionally, we enrich our data by incorporating five synthetic images for each object, created using stable diffusion. We show two samples of this external visual-name memory, constructed using objects from LVIS in Fig.~\ref{sup_fig_lvis}.

\subsection{WHOOPS memory}
To illustrate the scalability of the external memory in \method{}, we expand it by integrating WHOOPS knowledge into the original external visual--name memory in Sec.~\textcolor{red}{5.2} and Sec.~\textcolor{red}{5.3} of the main paper. Specifically, we focus on objects that are mentioned in the answers of VQA annotations in the WHOOPS dataset because of their conciseness and emphasis on key objects. For each of these objects, we produce five synthetic images employing stable diffusion. Two examples from this augmented memory, featuring newly added object images and their corresponding names, are presented in Fig.~\ref{sup_fig_whoops}.

\section{Additional Results}
\label{sup_sec_anal}

\input{tables/sup_tab_nocap_test}

\subsection{Experiments on NoCaps test set}
We additionally assess our \method{} against SOTAs on the NoCaps test set, since we notice several other methods have also benchmarked their performance on this dataset. Note that, NoCaps test set does not have publicly accessible ground truth annotations. To obtain evaluation scores, we submitted our results to the NoCaps leaderboard.

\noindent
\textbf{Quantitative results.}
Tab.~\ref{tab:nocaptest} presents the quantitative results of our \method{} on the NoCaps test set. Our method outperforms all other SOTA models, both in heavyweight-training and lightweight-training categories, that have reported results on this dataset. Additionally, as a lightweight method, our approach achieves the 8th rank on the NoCaps leaderboard, only surpassed by specialized SOTAs such as CogVLM, which holds the 1st rank.

\begin{figure*}[ht!]
\centering\includegraphics[width=1.00\textwidth]{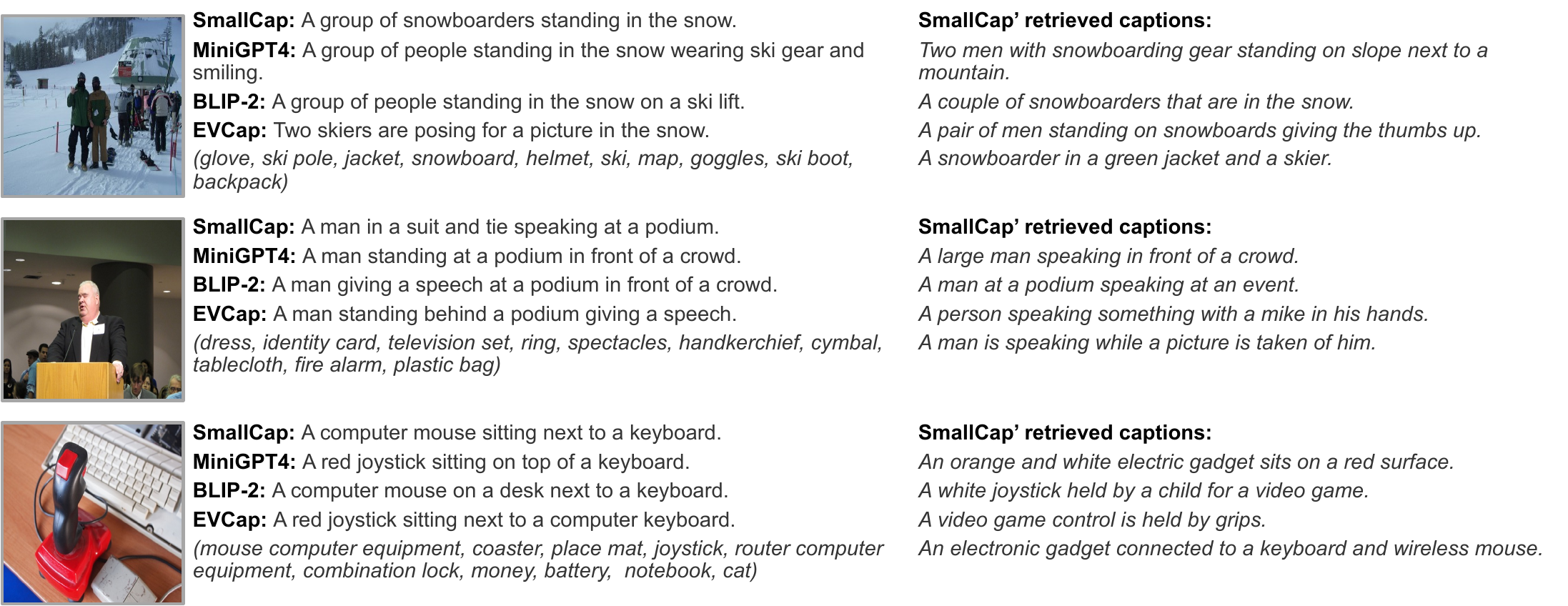}
\caption{Examples of captions generated by our \method{}, and three SOTAs on the NoCaps test set. We also list the retrieved object names by our \method{} (below the captions of \method) and retrieved captions by SmallCap (right side) in \textit{italics}.}
\label{sup_fig_nocaps_test}
\end{figure*}

\noindent
\textbf{Qualitative results.}
Fig.~\ref{sup_fig_nocaps_test} shows the captions generated by our \method{} alongside those from three SOTA methods on the NoCaps test set. It also includes the object names retrieved by \method{} and the captions retrieved by SmallCap. Consistent with the findings in the main paper, SmallCap tends to generate hallucinatory objects that are absent in the input images, such as \textit{``tie"} and \textit{``mouse"}. The same hallucinatory object \textit{``mouse"} is also found in the retrieved captions, indicating that SmallCap's diminished performance is largely due to its reliance on retrieved captions containing irrelevant information. In comparison, our \method{} demonstrates a performance on par with BLIP-2.

\begin{figure}[ht!]
\centering\includegraphics[width=0.43\textwidth]{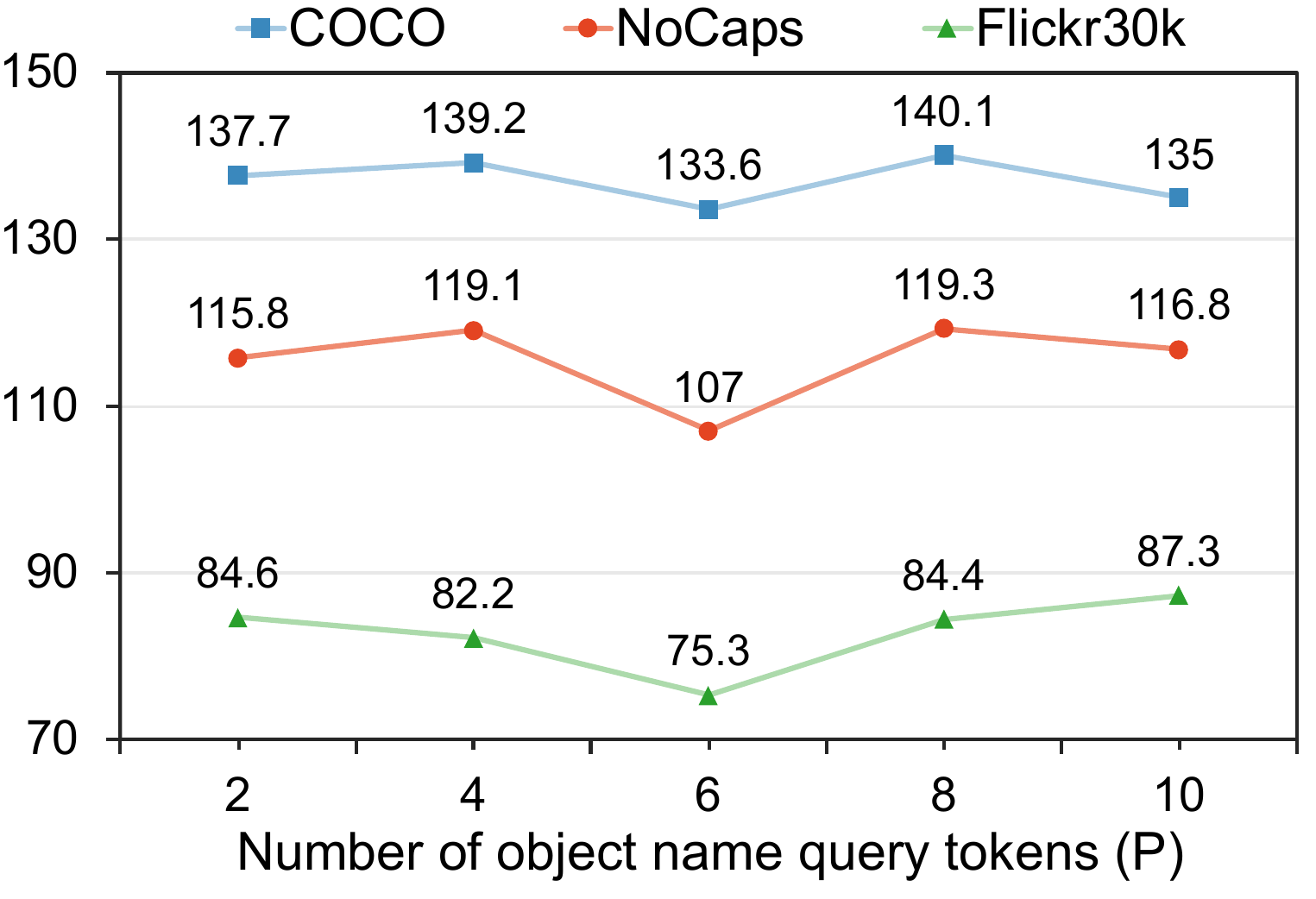}
\caption{CIDEr scores after training \method{} with the number of object name query tokens $P$ from 2 to 10. The results indicate that the performance is relatively optimal when $P$ is set to be 8.}
\label{sup_fig_tokens}
\end{figure}

\input{tables/sup_tab_time}

\subsection{Further analysis}

\noindent
\textbf{Comparison on training time and used GPUs.} 
Tab.~\ref{sup_tab:time} compares training time and the used GPUs of our \method{} with various SOTA models.
Due to the diversity of GPUs employed across different models, drawing a direct comparison is challenging. Nevertheless, it's evident that the training time for our \method{} is comparatively shorter than most models.

\noindent
\textbf{Number of object name query tokens.}  
We explore the impact of varying the number of retrieved object names $P$ (Sec.~\textcolor{red}{3.4} of the main paper) on \method{} in Fig.~\ref{sup_fig_tokens}. We train the model using different values of $P$, ranging from 2 to 10, and evaluate the performance under CIDEr on all three benchmarks. The results suggest that setting $P=8$ offers relatively optimal results.

\begin{figure*}[ht!]
\centering\includegraphics[width=1.00\textwidth]{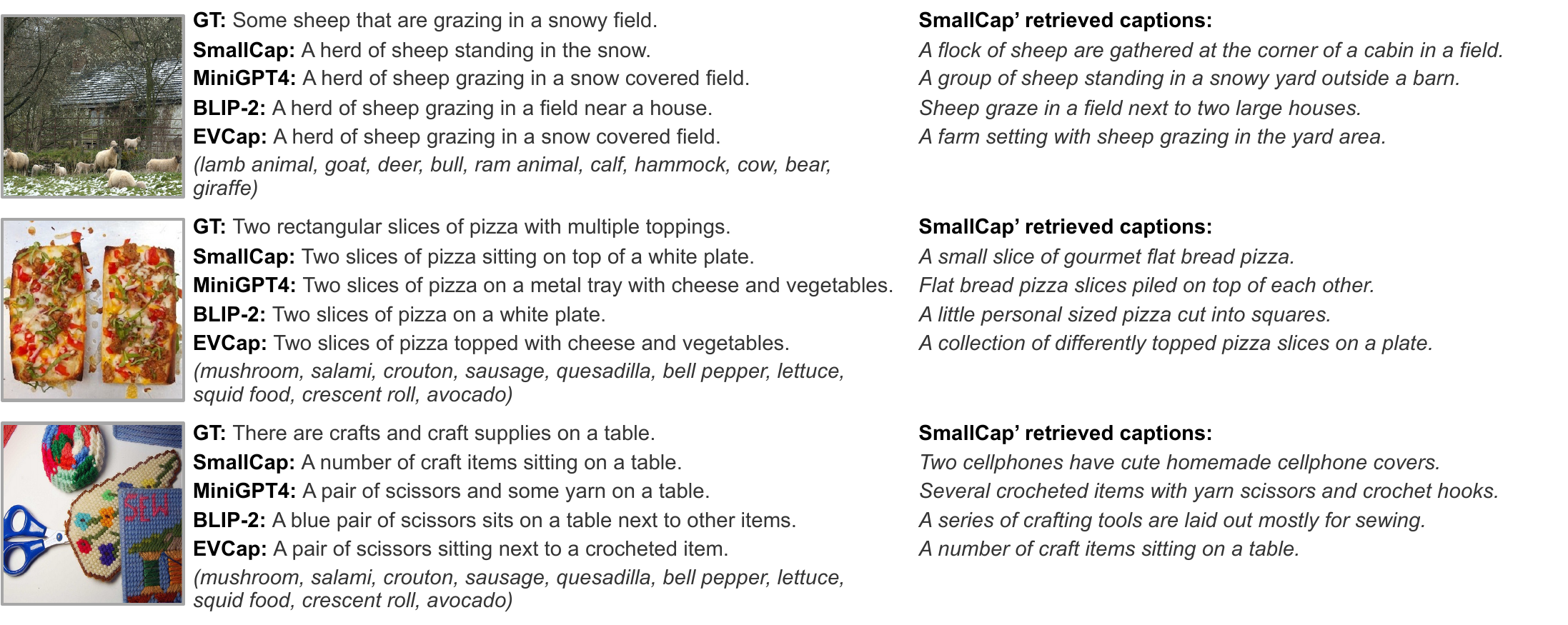}
\caption{Examples of captions generated by our \method{}, and three SOTAs on the COCO test set. We also list the retrieved object names by our \method{} (below the captions of \method) and retrieved captions by SmallCap (right side) in \textit{italics}.}
\label{sup_fig_coco}
\end{figure*}

\begin{figure*}[ht!]
\centering\includegraphics[width=1.00\textwidth]{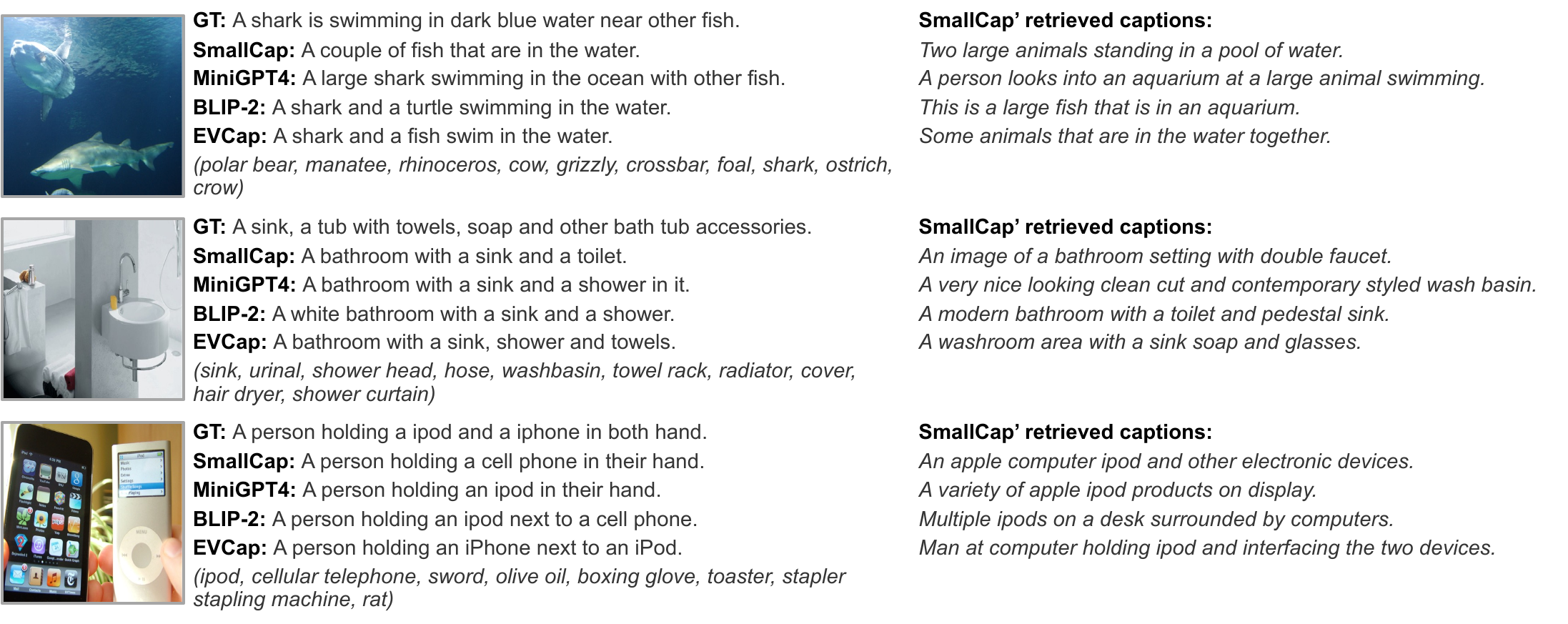}
\caption{Examples of captions generated by our \method{}, and three SOTAs on the NoCaps validation set. We also list the retrieved object names by our \method{} (below the captions of \method) and retrieved captions by SmallCap (right side) in \textit{italics}.}
\label{sup_fig_nocaps_val}
\end{figure*}

\subsection{More qualitative examples.}

\begin{figure*}[ht!]
\centering\includegraphics[width=1.00\textwidth]{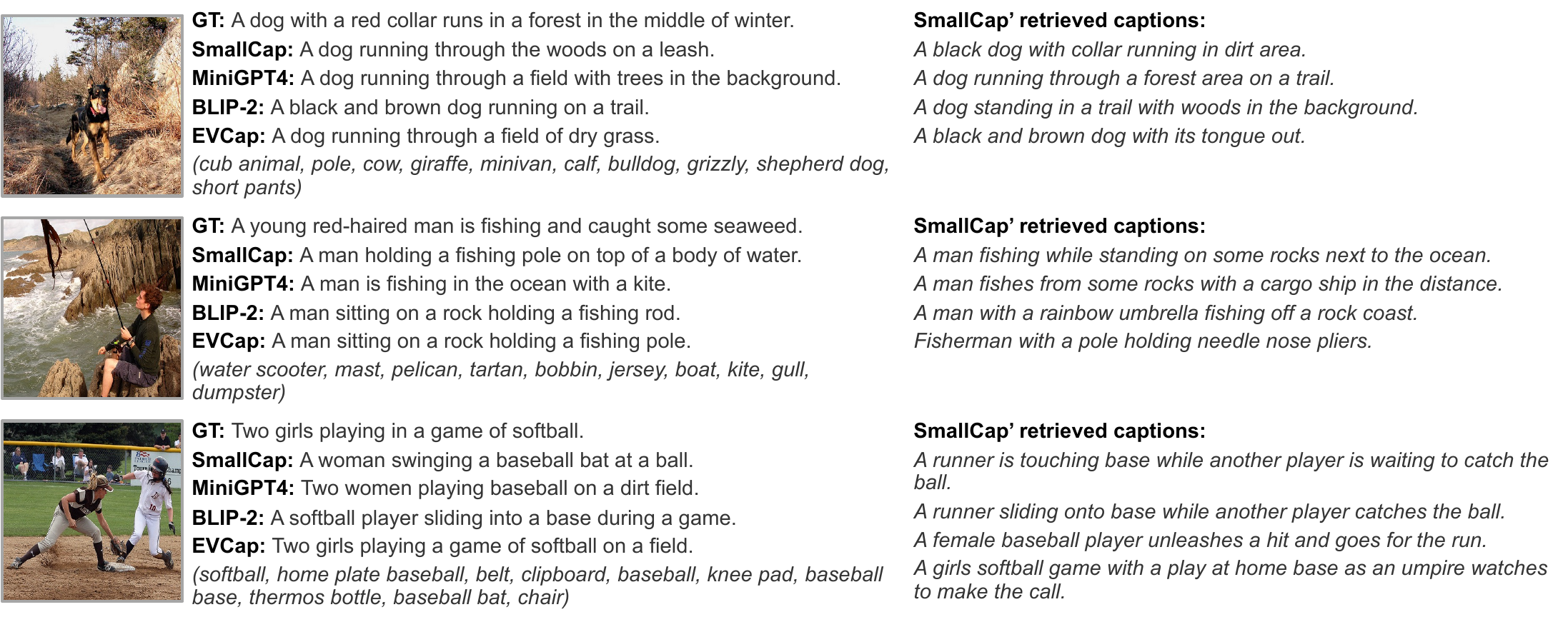}
\caption{Examples of captions generated by our \method{}, and three SOTAs on the Flickr30k test set. We also list the retrieved object names by our \method{} (below the captions of \method) and retrieved captions by SmallCap (right side) in \textit{italics}.}
\label{sup_fig_flickr30k_test}
\end{figure*}

More qualitative examples on the COCO test set, NoCaps validation set, and Flickr30k test set are shown in Fig.~\ref{sup_fig_coco}, Fig.~\ref{sup_fig_nocaps_val}, and Fig.~\ref{sup_fig_flickr30k_test}, respectively.

%% file: tables/sup_tab_nocap_test.tex
\begin{table}[tb]
\caption{Quantitative results under the CIDEr score on in-domain (In), near-domain (Near), out-domain (Out), and overall data of the NoCaps test set. * denotes using a \textbf{memory bank}. We note that our results on the test set are publicly submitted to Nocaps leader-board$^a$ (8th rank). Higher score is better. \colorbox{backblue1}{\textbf{Bold}} indicates the best results among compared methods, \colorbox{backblue}{normal} indicates the second best results.}
\label{tab:nocaptest}
\resizebox{\linewidth}{!}{
\begin{tabular}{lcccc}
\toprule[1pt]
\multirow{1}{*}{\textbf{Method}} 
& In & Near & Out & Overall\\ 
\midrule
\multicolumn{5}{l}{\textbf{Heavyweight-training models}} \\
VinVL~[\textcolor{cvprblue}{45}] & 93.8  & 89.0 & 66.1 & 85.5 \\ 
NOC-REK*~[\textcolor{cvprblue}{40}] & \colorbox{backblue}{100.0}  & \colorbox{backblue}{95.7} & 77.4 & \colorbox{backblue}{93.0} \\ 
OSCAR~[\textcolor{cvprblue}{26}] & 81.3  & 79.6 & 73.6 & 78.8 \\ 
\midrule
\multicolumn{5}{l}{\textbf{Lightweight-training models}} \\
SmallCap* \demph{$_\text{GPT2}$}~[\textcolor{cvprblue}{35}] & 87.9   & 84.6  & \colorbox{backblue}{84.4} & 85.0 \\ 
\rowcolor{Gray}
\method* \demph{$_\text{Vicuna-13B}$} & \colorbox{backblue1}{\textbf{114.9}}   & \colorbox{backblue1}{\textbf{117.0}}  & \colorbox{backblue1}{\textbf{117.1}} & \colorbox{backblue1}{\textbf{116.8}} \\ 
\midrule
\multicolumn{5}{l}{\demph{\textbf{Specialist SOTAs}}} \\
\demph{CogVLM \demph{$_\text{Vicuna-7B}$}~[\textcolor{cvprblue}{5}]} & \demph{--}  & \demph{--} & \demph{128.0} & \demph{126.4}\\
\demph{PaLI \demph{$_\text{mT5-XXL}$}~[\textcolor{cvprblue}{9}]} & \demph{--}  & \demph{--} & \demph{--} & \demph{124.4} \\
\demph{PaLI-X \demph{$_\text{UL2-32B}$}~[\textcolor{cvprblue}{8}]} & \demph{--}  & \demph{--} & \demph{--} & \demph{124.3} \\
\bottomrule[1pt]
\end{tabular}
}
\footnotesize{$^a$\url{https://eval.ai/web/challenges/challenge-page/355/leaderboard/1011}}
\end{table}

%% file: tables/sup_tab_time.tex
\begin{table}[tb]
{\centering
\caption{Comparison against SOTA methods on training time and used GPUs. * denotes using a \textbf{memory bank}.} 
\label{sup_tab:time}
\resizebox{\linewidth}{!}{
\begin{tabular}{l|cc}
\toprule[1pt]
\textbf{Method} & \textbf{Training time} & \textbf{GPUs} \\
\midrule
\multicolumn{3}{l}{\textbf{Heavyweight-training models}} \\
VinVL~[\textcolor{cvprblue}{45}] & --  & --   \\
AoANet+MA*~[\textcolor{cvprblue}{16}] & --  & --  \\
NOC-REK*~[\textcolor{cvprblue}{40}] & 8d & 2 RTX3090 \\
RCA-NOC*[\textcolor{cvprblue}{13}] & 1d & 8 A100 \\
ViECap \demph{$_\text{GPT2}$}[\textcolor{cvprblue}{15}] & --   & -- \\
InstructBLIP \demph{$_\text{Vicuna-13B}$}[\textcolor{cvprblue}{11}] & 1.5d  & 16 A100 \\
OSCAR~[\textcolor{cvprblue}{26}] & 74h  & 1 V100  \\
BLIP~[\textcolor{cvprblue}{24}]  & --  & 2 16-GPU nodes \\
BLIP-2 \demph{$_\text{FlanT5-XL}$}~[\textcolor{cvprblue}{25}]  & $\sim$9d  & 16 A100  \\
REVEAL* \demph{$_\text{T5}$}~[\textcolor{cvprblue}{20}]  & 5d  & 256 CloudTPUv4 chips  \\
\midrule
\multicolumn{3}{l}{\textbf{Lightweight-training models}} \\
MiniGPT4 \demph{$_\text{Vicuna-13B}$}~[\textcolor{cvprblue}{46}] & 10h & 4 A100 \\
SmallCap* \demph{$_\text{GPT2}$}~[\textcolor{cvprblue}{35}] & 8h  & 1 A100\\
ClipCap \demph{$_\text{GPT2}$}~[\textcolor{cvprblue}{29}]  & 6h  & 1 GTX1080  \\
\rowcolor{Gray}
\method* \demph{$_\text{Vicuna-13B}$} & 3h  & 4 A6000 \\
\midrule
\multicolumn{3}{l}{\demph{\textbf{Specialist SOTAs}}} \\
\demph{Qwen-VL \demph{$_\text{Qwen-7B}$}~[\textcolor{cvprblue}{5}]} & -- & -- \\
\demph{CogVLM \demph{$_\text{Vicuna-7B}$}~[\textcolor{cvprblue}{41}]} & \demph{1d}  & \demph{4096 A100} \\
\demph{PaLI \demph{$_\text{mT5-XXL}$}~[\textcolor{cvprblue}{9}]} & \demph{--}  & \demph{--}\\
\demph{PaLI-X \demph{$_\text{UL2-32B}$}~[\textcolor{cvprblue}{8}]} & \demph{--}  & \demph{--} \\
\bottomrule[1pt]
\end{tabular}
}
}
\end{table}